\documentclass[journal]{IEEEtai}
\let\labelindent\relax

\usepackage[colorlinks,urlcolor=blue,linkcolor=blue,citecolor=blue]{hyperref}

\usepackage{color,array}

\usepackage{graphicx}

\usepackage{booktabs}     
\usepackage{enumitem}
\usepackage{multirow}
\usepackage{tablefootnote}
\usepackage{graphicx}
\usepackage{subcaption}
\usepackage{bm}
\usepackage{algorithm}
\usepackage{algorithmic}
\usepackage{amsfonts} 
\usepackage{amsmath}
\usepackage{amsthm}
\usepackage{amssymb}
\usepackage{mathtools}
\usepackage{mathrsfs}
\usepackage{nicefrac} 
\usepackage{pifont}
\usepackage{tikz}
\usepackage[most]{tcolorbox}
\usepackage[section]{placeins}
\usepackage{cite}

\hypersetup{breaklinks}

\newtheorem{theorem}{Theorem}
\newtheorem{lemma}{Lemma}
\newtheorem{corollary}{Corollary}

\theoremstyle{definition}
\newtheorem{remark}{Remark}
\newtheorem{defn}{Definition}
\newtheorem{assumption}{Assumption}

\newcommand{\bs}{\boldsymbol}

\DeclarePairedDelimiter{\floor}{\lfloor}{\rfloor}
\DeclarePairedDelimiter{\abs}{\lvert}{\rvert}
\DeclareMathOperator*{\argmax}{argmax}

\definecolor{artun}{rgb}{0.90, 0.50, 0.30}

\newcommand{\algrule}[1][.2pt]{\par\vskip.5\baselineskip\hrule height #1\par\vskip.5\baselineskip}

\begin{document}

\title{Federated Multi-armed Bandits Under Byzantine Attacks}

\author{
	\IEEEauthorblockN{Artun Saday, İlker Demirel, Yiğit Yıldırım, and Cem Tekin,~\IEEEmembership{Senior Member,~IEEE}}
    \thanks{©2025 IEEE. Personal use of this material is permitted. Permission from IEEE must be obtained for all other uses, in any current or future media, including reprinting/republishing this material for advertising or promotional purposes, creating new collective works, for resale or redistribution to servers or lists, or reuse of any copyrighted component of this work in other works.}
    \thanks{This work was supported in part by the Scientific and Technological Research Council of Türkiye under Grant 123E467; in part by the Turkish Academy of Sciences Distinguished Young Scientist Award Program (TÜBA-GEBİP-2023). ({\em Corresponding Author: Cem Tekin})}
	\thanks{A. Saday, Y. Yıldırım and C. Tekin are with the Department of Electrical and Electronics Engineering, Bilkent University, Ankara, Türkiye, 06800 (email: artun.saday@bilkent.edu.tr, yigit.yildirim@bilkent.edu.tr, cemtekin@ee.bilkent.edu.tr).}
    \thanks{İ. Demirel is with Computer Science \& Artificial Intelligence Laboratory, Massachusetts Institute of Technology, Cambridge, MA, 02139 (email: demirel@mit.edu).}
}

\markboth{}
{Saday \MakeLowercase{\textit{et al.}}: Federated Multi-armed Bandits Under Byzantine Attacks}

\maketitle

\begin{abstract}
Multi-armed bandits (MAB) is a sequential decision-making model in which the learner controls the trade-off between exploration and exploitation to maximize its cumulative reward. Federated multi-armed bandits (FMAB) is an emerging framework where a cohort of learners with heterogeneous local models play a MAB game and communicate their aggregated feedback to a server to learn a globally optimal arm. Two key hurdles in FMAB are communication-efficient learning and resilience to adversarial attacks. To address these issues, we study the FMAB problem in the presence of Byzantine clients who can send false model updates threatening the learning process. We analyze the sample complexity and the regret of $\beta$-optimal arm identification. We borrow tools from robust statistics and propose a median-of-means (MoM)-based online algorithm, Fed-MoM-UCB, to cope with Byzantine clients. In particular, we show that if the Byzantine clients constitute less than half of the cohort, the cumulative regret with respect to $\beta$-optimal arms is bounded over time with high probability, showcasing both communication efficiency and Byzantine resilience. We analyze the interplay between the algorithm parameters, a discernibility margin, regret, communication cost, and the arms' suboptimality gaps. We demonstrate Fed-MoM-UCB's effectiveness against the baselines in the presence of Byzantine attacks via experiments.
\end{abstract}

\begin{IEEEImpStatement}
Federated learning systems became ubiquitous in many real world applications, such as healthcare, recommender systems, and communications. A critical challenge in ensuring such systems' robustness and communication efficiency involves dealing with malicious agents that send falsified updates. Our work develops a principled approach to Byzantine robust federated sequential decision-making by proposing a new federated multi-armed bandit problem. It not only characterizes the fundamental difficulty of finding the optimal arms in the presence of Byzantine clients but also rigorously bounds the sample complexity and regret of federated sequential decision-making. Thanks to its strong theoretical guarantees, our algorithm significantly improves the trustworthiness of federated sequential decision-making systems, paving the way for secure and resilient federated learning applications. Our fundamental theoretical results on learnability and Byzantine resilience, together with comprehensive experimental evaluations of the proposed algorithms, pave the way for the widespread adoption of secure and robust federated learning systems in the real world. 
\end{IEEEImpStatement}


\section{Introduction} \label{sec:intro}

\IEEEPARstart{M}{ulti}-armed bandits (MAB) model sequential decision-making under uncertainty where the goal is to maximize the collected reward under noisy feedback over a sequence of rounds from $1$ up to $T$ \cite{sutton1998introduction, pilarski2021optimal, bedi2023regret}. In its simplest form, the learner interacts with $K$ arms with unknown reward distributions. At each round, the learner pulls an arm and collects its immediate reward. In order to maximize the collected reward, the learner should intricately balance information collection and utilization. This led to the development of a series of algorithms that encoded exploration and exploitation either explicitly or implicitly in their arm selection mechanisms. Examples of the former include explore-then-commit \cite{robbins1952some} and interleaved exploration-exploitation \cite{vakili2013deterministic} style algorithms. Examples of the latter include Thompson sampling \cite{thompson1933likelihood} and upper confidence bound \cite{auer2002finite} strategies. Instead of measuring how high the rewards of these algorithms are, researchers have adopted a more interpretable benchmark, namely, how well these algorithms perform with respect to the best possible sequence of arms. This performance metric is known as the regret. It is known that any reasonable algorithm that performs well under a wide range of reward distributions should suffer at least $\Omega(\log T)$ expected regret \cite{lai1985asymptotically}. Remarkably, many variants of the aforementioned algorithms succeed in achieving $O(\log T)$ expected regret, confirming their time order optimality.

Another line of literature focused on best arm identification (BAI), where the goal is to identify the arm with the highest mean with fixed confidence in as few sequential evaluations as possible \cite{even2006action, karnin2013almost}. A relaxation of this is the $(\beta,\delta)$-{\em probably approximately correct} (PAC) BAI, where the goal is to identify an arm that is $\beta$ close to the optimal arm in terms of the mean reward with probability at least $1-\delta$ \cite{even2006action, garivier2021nonasymptotic}. 

In the literature, MAB models have been particularly instrumental in transforming recommender systems, where they address the challenge of providing personalized content to users by continuously learning from user interactions. A notable example is \cite{li2010news}, where a contextual bandit approach is employed to enhance news article recommendations. Other works highlighting the success of MAB algorithms in recommendation systems include \cite{chu2011contextual, kveton2015cascading}. Similarly, by skillfully matching tasks with the best-suited participants, MAB algorithms emerge as a powerful tool in crowd-sourcing  \cite{jain2014crowdsourcing, crowd_sourcing, gao2020crowdsourcing, elahi2022online, nee2018context, huang2023online}. Some of them smartly assign tasks outside of participants' usual expertise, uncovering hidden abilities and adjusting to skill changes. This method improves how tasks are assigned, making the process more efficient and successful. Recently, bandit algorithms have also been used for joint device activity detection and data decoding in random access with massive Internet-of-Things devices \cite{dong2022faster}.

Federated Learning (FL) has emerged as a novel paradigm in machine learning, motivated by the increasing need to train models on decentralized data while addressing privacy, security, and data governance concerns. Traditional centralized learning approaches, which require aggregating data in a single location, have become increasingly untenable due to strict data protection regulations like GDPR \cite{truong2021privacyGDPR}, the sheer volume of data generated at the edge, and the risks associated with data breaches. FL addresses these challenges by allowing the model to travel rather than the data, enabling a collaborative learning process across multiple devices or servers while keeping the data localized. This approach potentially mitigates privacy and security risks by minimizing data movement and also offers a partial solution to bandwidth constraints and latency issues associated with transferring large datasets. However, it introduces unique challenges, including managing model updates from heterogeneous models, ensuring robustness against adversarial attacks in decentralized environments, and dealing with the non-i.i.d. nature of real-world data across devices, which can significantly affect model convergence and performance \cite{li2020FL}.

MAB problems are inherently present in federated settings, one instance of which is item recommendation, where the central server aims to recommend the most popular item. However, due to privacy restrictions, the server may lack direct access to the user data required to estimate item popularity. To overcome this, the server can devise a method to interact with customers for feedback on popularity. Nonetheless, local item popularities among different customers can differ from the overall global popularity, leading to data heterogeneity. The server must infer the global item popularities from heterogeneous local bandit models, giving rise to a federated multi-armed bandit (FMAB) problem. In their pioneering work, \cite{shi2021federated} develop a framework to address this problem and formulate an algorithm called Fed2-UCB which utilizes a double UCB principle to adress both arm and client sampling. In the framework proposed by \cite{shi2021federated}, although clients make observations based on their individual, local arms, the reward mechanism is determined according to the global arm's performance. Subsequently, this concept was expanded by \cite{shi2021federatedper} in a later work to incorporate a personalized approach, where the reward structure is devised as a weighted sum of both local and global rewards. The study by \cite{fmab_privacy} addresses the issue of privacy within the FMAB framework, employing a differential privacy (DP) strategy. In all the scenarios discussed, communication cost is identified as a primary bottleneck.

Another key challenge in designing an FL system is to ensure that the global model is learned correctly even under falsified updates from the clients. Within this context, Byzantine robust federated learning is an important active research area. Notable works such as \cite{wan2022shielding} and \cite{wan2023four} try to recover from such attacks by rational client selection and the detection of Byzantine actors. Specifically, \cite{wan2022shielding} addresses this issue by treating the client selection as a MAB problem and introducing a graph-based method for Byzantine client detection. While prior research on FMAB has largely concentrated on model heterogeneity, the communication aspects, and resilience to potential adversarial behavior of clients as separate issues, there is a scarcity of studies that address these three aspects simultaneously. Our paper seeks to bridge this gap in the literature by exploring the three foundational pillars of FL, namely, communication efficiency, model heterogeneity, and Byzantine resilience within the context of FMAB.
In this work, we study the FMAB problem introduced in \cite{shi2021federated} under Byzantine attackers where a subset of the participating clients sends arbitrarily corrupted updates to the global server \cite{lamportbiz}. Our main contributions are as follows.
\begin{itemize}
	\item We propose a {\em median-of-means} (MoM) based algorithm, Fed-MoM-UCB, and a {\em trimmed mean} based algorithm, Fed-Trim-UCB, to maintain robustness against the Byzantine clients (Section~\ref{sec:fedmomucb_algo}).
	\item We derive $(\beta, \delta)$-PAC sample complexity results and a high probability $O(1)$ upper bound on the cumulative regret for the Fed-MoM-UCB algorithm (Section~\ref{sec:regretanalysismain}). 
	\item We conduct detailed experiments to compare Fed-MoM-UCB's performance against Fed2-UCB in \cite{shi2021federated} and Fed-Trim-UCB under Byzantine attacks and to investigate the effects of different parameters on the performance (Section~\ref{sec:experimentsmain}).
\end{itemize}

\section{Related Work} \label{sec:related}

\begin{table*}[ht]
	\centering
	\caption{Comparison of different works on FMAB problems}
	\label{tab:comparison}
	\begin{tabular}{lccccc}
		\toprule
		\textbf{Algorithm (Study)} & \textbf{Non-i.i.d. Data} & \textbf{Communication Cost} & \textbf{Adversarial Robustness} & \textbf{Regret Bounds} & \textbf{Identifiability Results} \\
		\midrule
		Fed2-UCB \cite{shi2021federated} & \checkmark & \checkmark & $\times$ & $O(\log T)$ & $\times$ \\
        HetoFedBandit \cite{blaser2024federated} & \checkmark & \checkmark & $\times$ & $\widetilde{O}(d\sqrt{MNT})$ & $\times$ \\
		Byzantine-UCB \cite{cdc_byzantine} & $\times$ & \checkmark & \checkmark & $\widetilde{O}(T^{3/4})$ & $\times$ \\  
		Fed-SEL \cite{byzantine_fed} & Distinct Arm Sets & \checkmark & \checkmark & $\times$ & $(0, \delta)\text{-PAC}$ \\ 
        FedSupLinUCB \cite{fan2024federated}  & $\times$ & \checkmark & \checkmark & $\widetilde{O}(\sqrt{dT} + d C_p)$ & $\times$ \\
        Fed-MoM-UCB (Our Work) & \checkmark & \checkmark & \checkmark & $O(1)$ w. $1-\delta$ prob. & $(\beta, \delta)\text{-PAC}$ \\        
    		\bottomrule
	\end{tabular}
\end{table*}

\subsection{Federated Learning} 

In recent years, FL has experienced significant advancements, with particular focus on addressing key challenges such as privacy, communication costs, and data heterogeneity \cite{robust_survey, vahidian2023rethinking}. The work by \cite{eth_tsne} introduces MiTFed, an innovative framework that leverages blockchain technology and smart aggregation to enhance robustness against Distributed Denial of Service (DDoS) attacks, ensuring privacy and facilitating collaboration in a decentralized context. Another study by \cite{kobi_comm} tackles the issues of communication overhead and data heterogeneity—two major obstacles in FL as identified by \cite{bonawitz2019towards}. The authors propose the COTAF algorithm, which preserves the convergence properties of local Stochastic Gradient Descent (SGD) while accommodating heterogeneous data across users. Additionally, \cite{client_select} addresses data heterogeneity by quantifying the skewness of client data in a privacy-preserving manner, subsequently selecting clients with less skewed data using a bandit approach. Scalability of FL systems in edge networks is studied in \cite{ching2024totoro}, where a fully decentralized FL system is shown to achieve significant gains in scalability and adaptivity.

A crucial requirement for an FL system is its robustness against arbitrary disruptions, known as Byzantine failures \cite{krum, lamportbiz}. A single Byzantine client can significantly undermine a model's reliability, spotlighting the importance of robust FL in current research \cite{kairouz2021advances}. Various robust estimators have been proposed, from those based on geometric medians \cite{geo_m2, gm_tsp} to sophisticated algorithms that identify corrupted updates through pairwise distances, such as Krum \cite{krum} and Auror \cite{auror}. Solutions also include trimmed-mean based estimators \cite{bulyan, geo_med, fang2020local} and explorations into federated learning under noisy communication conditions \cite{noisy, amiri2020federated}. For an extensive survey on privacy and robustness in FL, \cite{robust_survey} provides detailed insights. 

Some other related recent works include \cite{dong2023byzantine} which investigate Byzantine-robust distributed online learning, showing that even with robust aggregation, Byzantine-robust online gradient descent incurs linear adversarial regret. However, in i.i.d. environments, they achieve sublinear stochastic regret using momentum. Other studies such as \cite{zhu2023byzantine, mitra2022collaborative} also provide sublinear regret results under i.i.d. environments, where \cite{zhu2023byzantine} utilizes a trimmed mean estimator to achieve robustness and \cite{mitra2022collaborative} uses tools from high-dimensional robust Gaussian mean estimation. \cite{dong2024defending} considers a different approach and propose a robust federated learning framework that integrates blockchain technology to defend against poisoning attacks by decentralizing the model aggregation process. They combine the blockchain technology with a majority voting system to validate updates.

\subsection{Federated Multi-armed Bandits}

FMAB problems share common challenges with FL, and several studies have sought to address various FL aspects within the FMAB framework. Table \ref{tab:comparison} provides a comparison of different FMAB studies, outlining the specific challenges each addresses. Notably, \cite{shi2021federated} was among the first to introduce the FMAB framework, with a focus on data heterogeneity and communication costs. They achieve a regret bound of $O(\log(T))$. Another significant work is by \cite{cdc_byzantine}, which explores a federated linear bandit scenario involving Byzantine clients, albeit without considering data heterogeneity. Their main objective is to enhance the performance of each uncorrupted client by minimizing a metric they refer to as robust regret, using a robust geometric median aggregation. They manage to achieve a sublinear $\widetilde{O}(T^{3/4})$ regret, without data heterogeneity. \cite{blaser2024federated} also explores the federated linear bandit setting, but focuses on data heterogeneity, rather than Byzantine robustness. Their algorithm introduces a clustering approach that enables collaboration among clients with similar reward models. They obtain a regret bound of $\tilde{O}(d\sqrt{MNT})$ where $d$ is the dimension of the action space, $N$ is the number of clusters and $M$ is the number of clients. Their approach offers a fully personalized solution, positioning it at the other end of the spectrum compared to \cite{shi2021federated}.
A related work \cite{byzantine_fed} addresses the FMAB problem with a focus on both data heterogeneity and Byzantine robustness. Unlike our approach and that of \cite{blaser2024federated}, their concept of data heterogeneity is based on varying arm availability across clients, rather than differences in arm means. This study specifically targets the best arm identification problem, achieving high-probability identification guarantees. Their algorithm combines a local successive elimination phase with a global best arm identification phase. To ensure Byzantine robustness, they employ a majority voting mechanism with trimming to filter out adversarial influences.
Recently, global reward maximization by utilizing partial distributed feedback is studied by \cite{li2024distributed} within the context of linear bandits, while also aiming differential privacy.
Another study by \cite{fan2024federated} investigates the federated linear bandits problem and presents an algorithm that employs layered successive screening. They separately address the scenarios of heteroscedastic noise and adversarial attacks, providing a regret bound that scales linearly with the corruption budget in the adversarial setting. They achieve a regret bound of $\widetilde{O}(\sqrt{dT} + d C_p)$ where $d$ is the dimension of the action space and $C_p$ is the total corruption budget of the adversary.
Each study mentioned adopts a distinct approach, combining various aspects. In our work, we address both $(\beta,\delta)$-PAC BAI and regret minimization amidst heterogeneous data and Byzantine attacks, while also factoring in communication costs.

\begin{figure}
    \centering
    \includegraphics[width=1\linewidth]{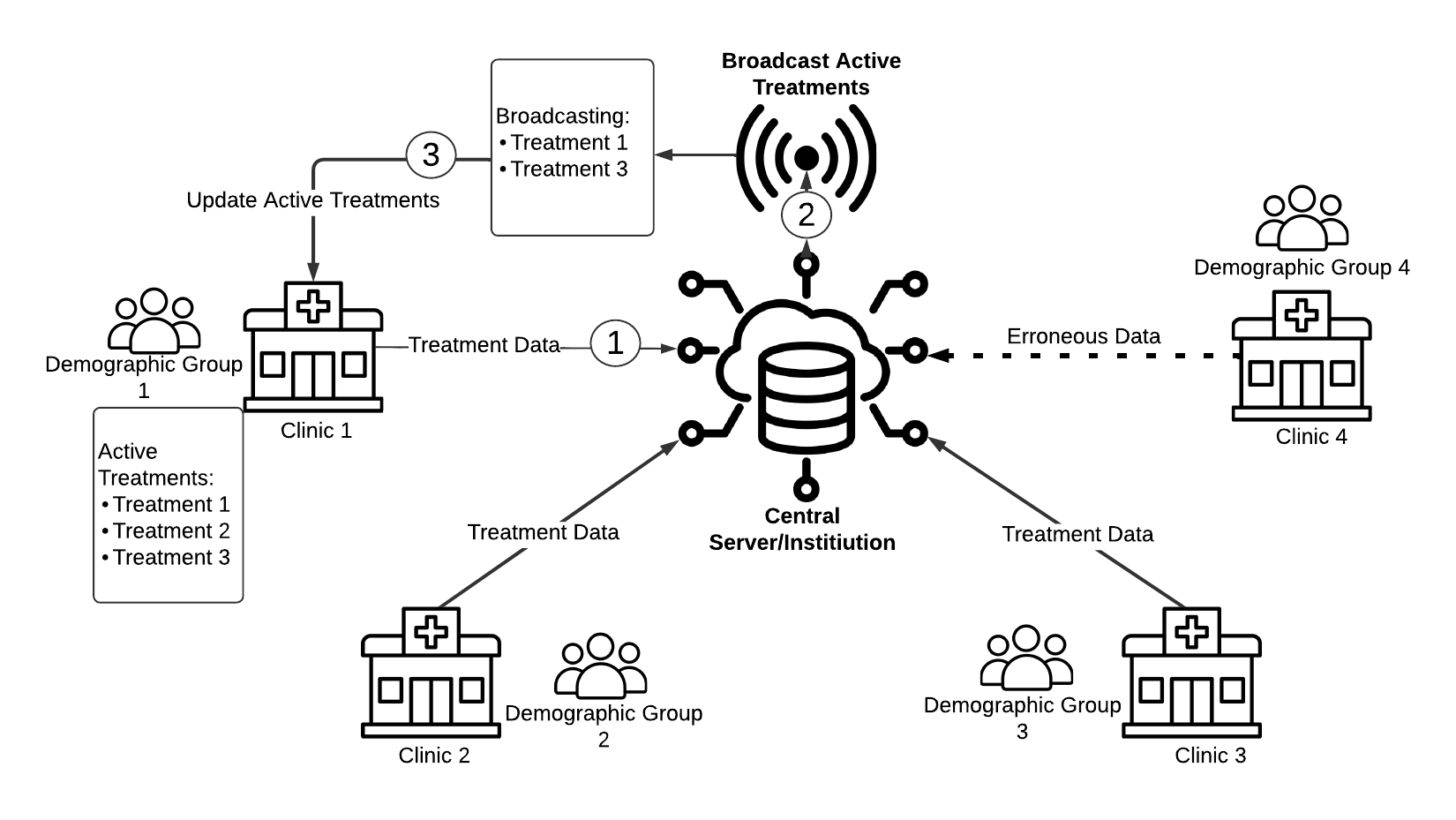}
    \caption{Illustration of the motivating example.}
    \label{fig:enter-label}
\end{figure}

\subsection{Motivating Example} \label{sec:motivating} 

Consider the problem of selecting effective treatments across various clinics or regions, coordinated by a central medical research institution as shown in Fig.~\ref{fig:enter-label}. Here, different treatment options can be represented as arms in a MAB problem. The problem is inherently federated, as data is collected from multiple clinics, with privacy regulations limiting the extent to which patient data can be shared. Additionally, model heterogeneity arises due to diverse patient demographics across clinics, leading to variations in treatment outcomes. Robustness against Byzantine clients is essential in this context, as erroneous data—whether from equipment malfunction or deliberate tampering—could result in the selection of a suboptimal or even harmful treatment. Instances of clinical malpractice in data handling are well-documented in the literature \cite{george2015data}.

In times of crisis, such as during a novel pandemic, quickly identifying a satisfactory treatment option is crucial to improving public health, framing this as a \((\beta, \delta)\)-PAC best arm identification problem. At the same time, minimizing regret is also important, as suboptimal treatments can have negative consequences for patient health. Motivated by these challenges, our study investigates the Federated Multi-Armed Bandit framework under Byzantine attacks, with a dual emphasis on best arm identification (BAI) and regret minimization.

\section{Problem Formulation} \label{sec:probformain}

\subsection{Federated MAB Setup}

We consider a federated learning system with a server and a pool of clients that can be recruited by the server. There are $K$ arms indexed by the set $[K] \coloneqq \{1,2,...,K\}$. Arm $k$ has an unknown {\em global mean reward} $\mu_k$. We call $\bs{\mu} := (\mu_k)_{k \in [K]}$ the {\em global model}. We denote the {\em global optimal arm} by $k^* \coloneqq \argmax_{k \in [K]} \mu_k$ and assume that it is unique. Clients cannot directly interact with the global model, but instead, they interact with their {\em local models}, which are stochastic realizations of the global model. In the local model of client $m$, the {\em local mean reward} of arm $k$ is generated as $\mu_{m,k} =  \mu_k + \xi_{m,k}$, where $\xi_{m,k}$ is the $\sigma_c$-subgaussian {\em model noise}. For another illustrative example of the model noise, consider a recommender system in which the server's objective is to pinpoint the item enjoying the highest global popularity, aiming to maximize profits. Due to the server's lack of direct access to global popularities, it relies on feedback from selected clients. Nonetheless, the popularities observed by clients at a local level may diverge from the global trends. The random variable $\xi_{m,k}$ represents this discrepancy. We assume that $\sigma_c$ is known by the server.

In FMAB, each client $m$ faces a stochastic $K$-armed bandit problem. In particular, playing an arm $k \in [K]$ in round $t$ yields the local reward
\begin{align} 
	X_{m,k} (t) = \mu_{m,k} + \zeta_{m,t} = \mu_k + \xi_{m,k} + \zeta_{m,t} ~. \label{eqn:reward_eqn}
\end{align}
In the above display, $\zeta_{m,t}$ is the {\em sampling noise} that is drawn from a fixed $\sigma$-subgaussian distribution independent of other clients and rounds. We assume that $\sigma$ is known by the server. 
Given $\beta>0$ and $\delta \in (0,1)$, the server and the clients' main joint goal is to identify an arm whose global mean is within $\beta$ of the global optimal arm with at least $1-\delta$ probability. In the bandit literature, this is also known as $(\beta,\delta)$-PAC BAI \cite{even2006action}.
Let $\Delta_k : = \mu_{k^*} - \mu_{k}$ denote the suboptimality gap for arm $k$. For $\beta \geq 0$, let $\mathcal{K}_{\beta} := \{ k \in [K] ~|~ \Delta_k \leq \beta \}$ denote the set of $\beta$-optimal arms. Let $\neg \mathcal{K}_{\beta} \coloneqq [K] \setminus \mathcal{K}_{\beta}$ denote the set of $\beta$-suboptimal arms. We define our objective as finding an arm $k \in \mathcal{K}_{\beta}$. Note that when $\beta < \min_{k \in [K] \setminus \{k^*\}} \Delta_k$, this coincides with finding the optimal arm.

While $\zeta_{m,t}$ is sampled at each round $t$, $\xi_{m,k}$ is sampled only once at the beginning, and $\mu_{m,k}$ is fixed from thereon. Equivalently, we can view $X_{m,k}(t)$ as a sample from an unknown distribution $\nu_{m,k}$. We call $\bs{\nu}_m := (\nu_{m,k})_{k \in [K]}$ the {\em local model} of client $m$. For different clients $m$ and $n$, we have $\mu_{m,k} \neq \mu_{n,k}$ in general. Since the local mean rewards $\mu_{m,k}$ are different from the global mean rewards $\mu_k$, a client $m$ may not agree with the global optimal arm, i.e., $k^* \notin \argmax_{k \in [K]} \mu_{m,k}$. In such a case, it is a hopeless task to identify the global optimal arm by relying on the local observations of a single client. 
However, the structure of the model noise suggest that by recruiting many clients, and averaging over their samples, one should be able to get an accurate estimate of the global mean rewards. This idea is exploited in \cite{shi2021federated}, where the interaction between the server and the clients proceeds in phases indexed by $p \geq 1$. At the end of each phase, clients send updates to the server, denoted by $U_{m,k}(p)$, which are estimates of the local mean rewards of arms. The server, then aggregates these updates to discard suboptimal arms and informs the clients. After this, the clients refine their updates by sampling the remaining arms in the next phase. This interaction continues until $k^*$ is identified. 

\subsection{Byzantine Clients}\label{sec:Byzantineclients}

At the end of phase $p$, an {\em honest client} $m$ sends the update $U_{m,k} (p) = \widehat{\mu}_{m,k}(p)$ for arm $k$, where  $\widehat{\mu}_{m,k} (p)$ is the sample mean of observations for arm $k$ collected by client $m$ up to phase $p$. On the other hand, a {\em Byzantine client} can send any $U_{m,k}(p)$, including $U_{m,k} (p)$ tailored to make the server's algorithm fail. Such malicious updates make identification of the global optimal arm challenging. Let $\mathcal{M}$ be the set of clients recruited by the server, and $\mathcal{M}^{\text{mal}} \subset \mathcal{M}$ be the set of Byzantine clients. Cardinalities of these sets are denoted by $M := \abs{\mathcal{M}}$ and $M^{\text{mal}} := \abs{\mathcal{M}^{\text{mal}}}$. The server and the honest clients do not know $\mathcal{M}^{\text{mal}}$. 

Even in the presence of a single Byzantine client, it may be impossible to identify the optimal arm using Fed2-UCB algorithm in \cite{shi2021federated} by naively relying on the sample means, no matter how many clients the server recruits. To see this, consider a single Byzantine client $m$ who sends the following updates as sample means for arm rewards to the server at the end of a phase $p$:
\begin{align}\label{eqn:cases_ma1_cli}
	U_{m,k} (p) = \begin{dcases}
		\widehat{\mu}_{m,k} (p) &\text{ if } k \neq k^* \\
		-\infty &\text{ if } k = k^*
	\end{dcases} 
\end{align}
Since Fed2-UCB calculates an average over individual clients' updates for an arm by the end of each phase, \eqref{eqn:cases_ma1_cli} will cause the global estimate for the optimal arm, $\overline{\mu}_{k^*}(p)$, to go to $-\infty$, rendering it impossible for Fed2-UCB to identify the global optimal arm. Similarly, a Byzantine client can favor any suboptimal arm to make it seem like the optimal arm. To conclude, one needs a robust federated learning algorithm that offers protection against Byzantine clients. 

In our study, we focus on the challenge of identifying $\beta$-optimal arms, namely, those arms residing within a $\beta$-proximity of the global optimal arm. To address the robustness concern, we introduce a {\em median-of-means} (MoM) aggregator for the server, detailed further in Section \ref{sec:fedmomucb_algo}. It will be impossible to learn $\beta$-optimal arms when all recruited clients are Byzantine. In order to obtain meaningful results, we make the following assumption on the ratio of the Byzantine clients. This assumption is standard in the robust statistics literature, where it is assumed that an upper bound on the fraction of corrupted samples is known \cite{altschuler2019best, trimmed, lafourge, byzantine_fed}. Similar assumptions are common in the federated learning literature addressing Byzantine attacks. For instance, \cite{byzantine_fed, dong2023byzantine, cdc_byzantine, mitra2022collaborative} assume an estimated upper bound on the number of Byzantine clients, while \cite{fan2024federated} assumes a bounded total corruption.

\begin{assumption}\label{ass:maliciousratio}
	At most $0 \leq \lambda < 0.5$ fraction of the clients recruited by the server are Byzantine. The server knows $\lambda$.
\end{assumption}

\subsection{Regret}

In addition to $\beta$-optimal arm identification, we also want our algorithm to avoid sampling $\beta$-suboptimal arms unnecessarily, until all $\beta$-suboptimal arms  are eliminated. We measure the performance of the interaction protocol between the server and the clients by introducing the notion of {\em robust federated regret}. Let $I_t^m$ denote the arm selected by client $m$ at round $t$. We denote by $\mathcal{M}_{t}^{\beta} \subseteq \mathcal{M}$ the set of clients who plays an arm $k \in \neg \mathcal{K}_{\beta}$ at round $t$. Robust federated regret defined below accounts for the cumulative loss of all clients with respect to $\beta$-optimal arms by considering the communication cost between the server and clients.
\begin{align}
	R^{\beta}_T \coloneqq \sum_{t=1}^T \sum_{m \in \mathcal{M}_{t}^{\beta}} (\Delta_{I_t^m} - \beta)_+ + C M N_T~,\label{eqn:regret2}
\end{align}
where $(\cdot)_{+} \coloneqq \max\{ \cdot, 0 \}$. $C > 0$ is a constant which denotes the cost incurred in a single communication round between the server and a client, and $N_T$ represents the total number of communication rounds by the end of round $T$. 

\section{Fed-MoM-UCB Algorithm} \label{sec:fedmomucb_algo}

We propose Fed-MoM-UCB: {\em Federated Median-of-Means Upper Confidence Bound} algorithm for $\beta$-optimal arm identification under Byzantine clients. In addition, Fed-MoM-UCB will also guarantee small robust federated regret. It is a distributed algorithm that consists two pieces: one piece running independently on each client, and one piece running on the server. In Fed-MoM-UCB, the clients roughly follow  the same arm sampling rule as in Fed2-UCB in \cite{shi2021federated}. However, the server employs an entirely different MoM-based robust estimator for global arm outcomes, which constitutes the key component of Fed-MoM-UCB. The server communicates with the clients at the end of every phase. Each phase lasts for a certain number of rounds communicated to the clients by the server. Next, we explain in detail client and server sides of our distributed algorithm. 

\subsection{Client Side} 
At the beginning of a phase $p$, a client $m$ receives the set of active arms ${\mathcal{A}}_p \subseteq [K]$ from the server. Then, it samples each arm in ${\mathcal{A}}_p$ to accumulate a total of $s(p)$ samples from each arm by the end phase $p$. Let $l(p) := s(p) - s(p-1)$ denote the number of samples obtained from an arm $k \in {\mathcal{A}}_p$ in phase $p$. The {\em sampling budget} $s(p) \in \mathbb{Z}_+$ increases with $p$, and is communicated to the clients by the server. At the end of a phase $p$, an honest client $m$ forms its local update vector as
%
	$\bm{U}_m (p) = ( \widehat{\mu}_{m,k} (p) )_{k \in {\mathcal{A}}_p}$,
%
where $\widehat{\mu}_{m,k} (p)$ is the sample mean of the observations for arm $k$ collected by client $m$ by the end of phase $p$. On the other hand, a Byzantine client $m$ can send an arbitrary $\bm{U}_m (p)$ to the server (see, e.g., \eqref{eqn:cases_ma1_cli}).

\begin{algorithm}[t!] 
	\caption{Fed-MoM-UCB (Server side)}
	\begin{algorithmic}[1]
		\label{alg:algo1}
		\STATE \textbf{Inputs}: $K$, $\delta$, $\beta$, $\lambda$, $\alpha(\lambda)$, $\sigma$, $\sigma_c$, $s(p)$, $p \geq 1$
		\algrule[2pt]
        \STATE Set number of clients in a group $B = \lfloor 1 / \alpha(\lambda) \rfloor$\\
		\STATE Set number of clients to recruit $M$ as in Lemma \ref{lemma:confidence_bound}
		\STATE Recruit $M$ clients to form the client set ${\cal M}$
		\STATE Set $t=0$, $p=1$, ${\cal A}_1 = [K]$, $E_0 = +\infty$
		\WHILE{$E_{p-1} > \beta/4$ and $\abs{{\cal A}_p} > 1$} 
		\STATE Ask each client $m \in {\mathcal{M}}$ to collect a total of $s(p)$ observations from each active arm $k \in \mathcal{A}_p$ \\    
		\STATE Obtain local updates $\bm{U}_m (p)$ from each client $m \in {\mathcal{M}}$ \\
		\STATE Divide the clients into $G = M/B$ groups \\
		\STATE Calculate group means $U^i_k(p)$ for $i \in [G]$ and $k \in {\cal A}_p$ as in \eqref{eqn:intermean} \\
		\STATE Form the MoM estimates $\overline{U}_k(p)$ for $k \in {\cal A}_p$ as in \eqref{eqn:globalupdate} \\
		\STATE Form the active arm set ${\cal A}_{p+1}$ for next phase as in \eqref{eqn:s3ae} \\
		\STATE $t \leftarrow t + \big(s(p) - s(p-1) \big) |{\mathcal{A}}_p|$  \\ 
		\STATE $p \leftarrow p + 1$ 
		\ENDWHILE
		\STATE (Termination) clients continue pulling the arms in ${\cal A}_p$
	\end{algorithmic}
\end{algorithm}

\subsection{Server Side}

In this subsection, we describe details of the server's algorithm given in Algorithm \ref{alg:algo1}. We assume that the server is given a suboptimality level $\beta >0$. Its goal is to identify a $\beta$-optimal arm as soon as possible. Due to the {\em model noise}, it is impossible to learn a $\beta$-optimal arm by relying on reward samples from a single client. On the other hand, the server can accurately estimate the global mean outcomes ($\mu_k$s) by averaging over samples of many clients. Therefore, at the beginning, the server recruits $M$ clients. The optimal choice of $M$ is discussed in Section~\ref{ssec:momconc} (see Lemma \ref{lemma:confidence_bound}). However, as detailed in Section~\ref{sec:Byzantineclients}, simply averaging the local updates of all clients does not work, as even a single Byzantine client can negatively influence the estimation of $\mu_k$s. In order to reduce the influence of the Byzantine clients in the learning process, the server randomly divides the clients into $G$ groups, where each group consists $B$ clients.\footnote{$M$ is chosen to be an integer multiple of $B$, which ensures that each group has the same number of clients and $G$ is well defined.} Next, we detail how dividing the clients into groups protects against Byzantine clients and allows a sufficient amount of model averaging to reduce the effect of the modeling noise on the estimation process.

\begin{figure}[t]
	\centering
	\includegraphics[width = \linewidth]{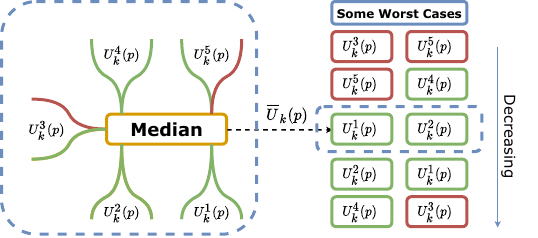}
	\caption{Ten clients where two are Byzantine (red lines), and they perform outlier attacks. Fed-MoM-UCB first divides the clients into five groups. Two example groupings are shown on the right. In the worst case, the Byzantine clients are assigned to different groups and saturate their groups' local updates. The saturation will not affect the median estimate.}
	\label{fig:algsch}
\end{figure}

At the end of phase $p$, the server collects the local updates $\bs{U}_m(p)$ from each client $m \in {\cal M}$. For $i \in [G]$, let ${\mathcal{B}}_{i,p}$ represent the set of clients in group $i$ in phase $p$. To reduce the effect of the model noise, the {\em group mean} of arm $k \in {\mathcal{A}}_p$ for group $i$ is calculated as
\begin{align} 
	U^i_k(p) &= \dfrac{1}{B} \sum_{ m \in {\mathcal{B}}_{i,p} } \widehat{\mu}_{m,k}(p) ~. \label{eqn:intermean}
\end{align}
Note that a Byzantine client in ${\mathcal{B}}_{i,p}$ can alter $U^i_k(p)$ arbitrarily. To protect against such clients, the server merges the group means for arm outcomes by using a {\em median-of-means} (MoM) estimator to form estimates for the global arm rewards as $\overline{\bm{U}}(p) = \{\overline{U}_k(p) \}_{k \in {\mathcal{A}}_p}$, where
\begin{align}
	\overline{U}_k(p) =  \text{med}\big( \{U^i_k(p)\}_{i \in [G]} \big)~. \label{eqn:globalupdate}
\end{align} 
Since the local updates are mean arm outcomes, the resulting estimator is a {\em median-of-means-of-means} (MoMoM) estimator. We will prove that $\overline{U}_k(p)$ is a sufficiently accurate estimate of $\mu_k$ when at least half of the groups consist of only honest clients. %
Fig.~\ref{fig:algsch} provides an example where the Byzantine clients perform outlier attacks, which are discarded thanks to the grouping strategy.

While increasing the number of groups offer protection against Byzantine clients, we also want each group to contain as many clients as possible to decrease the effect of the model noise on the estimates. To control the number of groups $G$, we introduce the following generic mapping as in \cite{lafourge}.
\begin{defn} \label{def:alpha}
	Let $\alpha : [0 , 1/2] \rightarrow [0,1]$ be a mapping such that
		$\forall \lambda \in [0, 1/2),~ 2 \lambda < \alpha(\lambda) < 1$. Table~\ref{tab:tau} lists examples of $\alpha$.
\end{defn}

\begin{table}[h] 
	\centering
	\caption{Different choices $\alpha(\lambda)$ satisfying Definition~\ref{def:alpha}.}
	\label{tab:tau}
	\begin{tabular}{@{}llcccc@{}}
		\toprule
		Mapping           & $\rightarrow$ & Arithmetic       & Geometric          & Harmonic               & Polynomial             \\ \midrule
		$\alpha(\lambda)$ & $\rightarrow$ & $(1+2\lambda)/2$ & $\sqrt{2 \lambda}$ & $4\lambda/(1+2\lambda)$ & $\lambda(2.5-\lambda)$ \\ \bottomrule
	\end{tabular}
\end{table}

Given $\alpha$, the server sets $B = \floor{1 / \alpha(\lambda)}$, and $G = M/B$. After obtaining $\overline{\bs{U}}(p)$, it identifies the suboptimal arms as
\begin{align*}
	 {\mathcal{ S}}_p := \{ k \in {\mathcal{ A}}_p | \overline{U}_k(p) + E_p  < \max_{k \in {\mathcal{A}}_p} \overline{U}_k(p) - E_p \} ~.
\end{align*}
The {\em uncertainty level} $E_p \in \mathbb{R}_{+}$ is chosen to satisfy for a given {\em confidence parameter} $\delta \in (0,1)$
\begin{align} 
	\mathbb{P} ( |\overline{U}_k (p) - \mu_k| \leq E_p ,~\forall p \in [\Bar{p}], \forall k \in {\cal A}_p ) \geq 1-\delta ~,  \label{eqn:s3cb}
\end{align}
where $\Bar{p}$ represents the maximum number of phases after which the algorithm will surely terminate, whose exact value will be derived in the next section. When the event in \eqref{eqn:s3cb} holds, we have for any $k \in {\cal S}_p$, $\mu_k < \mu_{k^*}$. We provide detailed explanations on how to choose $E_p$ and $M$ to guarantee the high probability condition in \eqref{eqn:s3cb} in the next section. 
After the suboptimal arms are identified, they are eliminated, and the active arm set for the next phase is computed as
\begin{align}
	{\mathcal{A}}_{p+1} = {\mathcal{ A}}_p \setminus  {\mathcal{ S}}_p ~.  \label{eqn:s3ae}
\end{align}
This continues until either $|{\cal A}_p|=1$ or when the uncertainty level reaches $E_{p-1} \leq \beta/4$, which we call the {\em termination phase}. The latter is sufficient to declare the arms in ${\cal A}_p$ as $\beta$-optimal (see Lemma~\ref{lemma:termination}).

\begin{remark} \label{rem:altregret}
    Since our primary objective is to identify a set of $\beta$-optimal arms, our analysis concentrates on the algorithm's regret after it has completed the elimination of $\beta$-suboptimal arms and all communication rounds have concluded. We refer to this state as the \emph{termination phase} of our algorithm. In Section \ref{sec:regret}, we show that Fed-MoM-UCB successfully reaches the termination phase, and provide sample complexity and regret bounds at this point. More specifically, we denote the termination phase of our algorithm as $p_\tau$ and denote the regret at termination as
\begin{align*}
    \Tilde{R}^{\beta}_{\tau} := \sum_{p=1}^{p_{\tau}} \sum_{m=1}^M \sum_{i \in {\cal A}_p}  \sum_{t \in {\cal T}_{i,m,p}} (\Delta_{I_t^m} - \beta)_+ + C M p_{\tau} ~.
\end{align*}
where ${\cal T}_{i,m,p}$ is the set of rounds in which client $m$ plays arm $i$ in phase $p$ of Fed-MOM-UCB. $\Tilde{R}^{\beta}_{\tau}$ is equivalent to the formulation in \eqref{eqn:regret2}, when $T$ is taken to be the last round of the algorithm before termination.
\end{remark}

\section{Theoretical Analysis} \label{sec:regretanalysismain}

In this section, we analyze the sample complexity and the robust federated regret of Fed-MoM-UCB given in \eqref{eqn:regret2}. 

\subsection{Preliminary Technical Results} \label{ssec:momconc}

The following lemma presents a concentration bound for the MoM estimate of a global arm outcome $\mu_k$ in a phase~$p$.
\begin{lemma} \label{lemma:global}(MoM concentration bound).
	Fix $k \in [K]$. Let $B=\floor{1/\alpha(\lambda)}$, $G = M/B$, $\eta(\lambda) \coloneqq (\alpha(\lambda) - \lambda)/\alpha(\lambda)$, and $\rho := \sigma/\sqrt{s(p)} + \sigma_c$. Then, for all 
	\begin{align} \label{eqn:s4i1}
		&\gamma_p^2 \geq 4\left( \frac{\sigma}{\sqrt{s(p)}} + \sigma_c \right)^2 \left(\frac{4\eta(\lambda)-1}{2\eta(\lambda)-1} \frac{\log 2}{B}\right) , \text{we have} \\ 
		&\mathbb{P} ( |\overline{U}_k (p) - \mu_k| > \gamma_p ) \leq \exp \left( - \frac{2 \eta(\lambda) -1}{16 \rho^2} M \gamma_p^2\right)~. \label{eqn:lem1}
	\end{align}
\end{lemma}

The proof of Lemma~\ref{lemma:global} utilizes the fact that a mapping $\alpha(\lambda)$ as defined in Definition \ref{def:alpha}, ensures that at least half of the groups consists solely of honest clients. The complete proof in detail is provided in the Appendix. Lemma~\ref{lemma:global} offers guidance on selecting $E_p$ in \eqref{eqn:s3cb} and $M$, while also highlighting a limitation: it is not possible to achieve probability guarantees for an arbitrarily small uncertainty level $\gamma_p$ in \eqref{eqn:lem1}. The smallest uncertainty level under which \eqref{eqn:lem1} holds is equal to the square root of the term that appears on the r.h.s. of \eqref{eqn:s4i1}. Therefore, we set the uncertainty level $E_p$ as
\begin{align} \label{eqn:s4i2}
	E_p = 2\left( \frac{\sigma}{\sqrt{s(p)}} + \sigma_c \right) \left(\frac{4\eta(\lambda)-1}{2\eta(\lambda)-1} \frac{\log 2}{B}\right)^{1/2}~.
\end{align}
In the above display, the term $\sigma/\sqrt{s(p)}$ accounts for the sampling noise while the term $\sigma_c$ accounts for the model noise. 

\begin{remark} \label{rmr:error}
	For a fixed mapping $\alpha(\lambda)$, the group size $B = \floor{1/\alpha(\lambda)}$ does not increase with $p$. This is required to ensure that at each phase $p$, at least half of the groups consist of all honest clients. As the number of clients in a group cannot increase with $p$, the uncertainty emerging from the {\em model noise} within a group does not decrease over phases. This effects global estimates since the server uses a median-of-means estimator. This is why we recruit all the clients once at the beginning. The reason for dividing algorithm into phases is to avoid oversampling suboptimal arms to minimize regret. 
\end{remark}

We note that $\beta$ cannot be arbitrarily small, as the lower bound of $\gamma_p$ in Lemma \ref{lemma:global} cannot be made arbitrarily small as discussed in Remark \ref{rmr:error}. Looking at $E_p$ in \eqref{eqn:s4i2}, no matter how large $s(p)$ is, the uncertainty level remains $O(\sigma_c)$.
Therefore, we introduce a lower bound for an achievable $\beta$, named the {\em discernibility margin}, defined as
\begin{align}
		\omega \coloneqq 8\sigma_c \left( \frac{4\eta(\lambda)-1}{2\eta(\lambda)-1} \frac{\log 2}{B} \right)^{1/2} ~. \label{eqn:omega}
\end{align}
In particular, we will show that any arm $k$ with suboptimality gap $\Delta_k > \omega$, will be eliminated after at most $O( (\Delta_k - \omega)^{-2} )$ rounds. Fed-MOM-UCB does not have elimination guarantees for arms with suboptimality gap smaller than $\omega$. The main results in hold under the following assumption.
\begin{assumption} \label{ass:discern}
$\beta > \omega$. 
\end{assumption}

 \begin{remark}\label{rem:marginremark}
     For all $\alpha(\lambda)$ in Table~\ref{tab:tau} except arithmetic mapping, $\lim_{\lambda \rightarrow 0} \omega = 0$. 
 \end{remark}
Let $\epsilon \coloneqq \beta - \omega$, and let $\Lambda_1 \coloneqq \frac{64 (4\eta(\lambda) - 1) \log 2}{  (2\eta(\lambda) - 1) B}$, $\Lambda_2 \coloneqq \frac{4}{(4 \eta(\lambda) - 1) \log2}$ be terms that only depend on $\lambda$. The next lemma gives an upper bound, denoted $\Bar{p}$, on the number of phases until termination.

\begin{lemma}(Phases until termination) \label{corr:pt}
\begin{align*}
    \Bar{p} = \min \left\{ p \in \mathbb{N}_{+} \bigg| s(p) \geq \frac{\Lambda_1 \sigma^2}{\epsilon^2} \right\} ~.
\end{align*}
\end{lemma}
	\begin{proof}
	By the definitions of $\omega$ in \eqref{eqn:omega} and $E_p$ in \eqref{eqn:s4i2}, we have
	\begin{align} \label{eqn:fli11pt}
		E_p = \frac{\omega}{4} + \frac{2 \sigma}{\sqrt{s(p)}}  \left(\frac{4\eta(\lambda)-1}{2\eta(\lambda)-1} \frac{\log 2}{B}\right)^{1/2}~.
	\end{align}
	 Fed-MoM-UCB terminates when $E_p \leq \beta/4 = (\omega + \epsilon)/4$. Thus $\bar{p}$ is the smallest integer satisfying
	\begin{align} 
		& \frac{2 \sigma}{\sqrt{s(\bar{p})}}  \left(\frac{4\eta(\lambda)-1}{2\eta(\lambda)-1} \frac{\log 2}{B}\right)^{1/2} \leq \frac{\epsilon}{4} 
		\Leftrightarrow s(\bar{p}) \geq \frac{\Lambda_1 \sigma^2}{\epsilon^2}  ~. \notag 
	\end{align}
    \end{proof}

Lemma \ref{corr:pt} is based on the fact that as the sampling budget $s(p)$ increases, the uncertainty level $E_p$ decreases and approaches $\omega/4$ as $p$ grows. Consequently, for $\beta > \omega$, there exists a phase $p$ at which the termination condition $E_p \leq \beta/4$ in Algorithm \ref{alg:algo1} is met. Next, we derive how many clients to recruit such that when $E_p$ is given as in \eqref{eqn:s4i2}, \eqref{eqn:s3cb} holds. Let
	\begin{equation} \label{eqn:good}
		{\cal E}_{g} \coloneqq \left\{ \abs{\overline{U}_k(p) - \mu_k} \leq E_p, \forall p \in [\Bar{p}], \forall k \in \mathcal{A}_p\right\}
	\end{equation}
 be the {\em good event} where the confidence bounds hold over all active arms and all phases $p \leq \Bar{p}$.

\begin{lemma}\label{lemma:confidence_bound}
Let $\delta \in (0,1)$ and $\alpha(\lambda)$ be given as in Definition~\ref{def:alpha}. When $M = \left\lceil \Lambda_2 \log(K \bar{p}/ \delta) \right\rceil B$ and $E_p$ is set as in \eqref{eqn:s4i2}, we have $\mathbb{P} ( {\cal E}_{g} ) \geq 1-\delta$. 
\end{lemma}

The proof of Lemma \ref{lemma:confidence_bound}, provided in the Appendix, combines the confidence bound \eqref{eqn:lem1} from Lemma \ref{lemma:global} with a union bound over all phases $p \in [\bar{p}]$. Then, the number of clients $M$ is chosen as the smallest value that ensures $P({\cal E}_g) \geq 1 - \delta$. Choosing a larger $M$ is unnecessary, as it does not reduce the discernibility margin, as discussed in Remark \ref{rmr:error}. Additionally, the fixed client selection scheme used in Fed-MoM-UCB is simpler to implement than schemes that increase the number of clients with each phase, as in \cite{shi2021federated}. In the rest of the paper, we always assume that $E_p$ is set as in \eqref{eqn:s4i2} and $M$ as in Lemma \ref{lemma:confidence_bound}.
\begin{remark} \label{rem:aboutM}
    Note that $M$ depends on $\bar{p}$ only through the $\log(\cdot)$ term. Since $\bar{p} \leq \frac{\Lambda_1}{\epsilon^2}$ for any choice of $l$, the number of clients we need to recruit only depends logarithmically on $1/\epsilon$.
\end{remark}

\subsection{$(\beta,\delta)$-PACness, Sample Complexity and Regret Analysis} \label{sec:regret}

The next lemma upper bounds the number of samples accumulated from an arm by any client until termination.

\begin{lemma}(Samples until termination) \label{corr:pmax}
		Let $p_{\tau}$ represent the phase after which Fed-MOM-UCB terminates. We have 
		\begin{align} \label{eqn:fli10}
			s(p_{\tau}) < \frac{\Lambda_1 \sigma^2}{\epsilon^2} + l(p_{\tau})~.
		\end{align}
\end{lemma}
\begin{proof}
This result directly follows from Lemma \ref{corr:pt}, noting that $p_{\tau} \leq \Bar{p}$, $s(p) = s(p-1) + l(p)$, and $\bar{p}$ is the smallest integer satisfying $s(\bar{p}) \geq \frac{\Lambda_1 \sigma^2}{\epsilon^2}$. 
\end{proof}

At the end of each phase $p$, Fed-MOM-UCB executes the arm elimination rule in \eqref{eqn:s3ae}. We first show that the globally optimal arm $k^*$ is guaranteed to always remain active.
\begin{lemma} \label{lemma:globalremains}
	Assume that the good event ${\cal E}_{g}$ in \eqref{eqn:good} holds. Then, $k^* \in {\cal A}_p$ for all $p \in [\Bar{p}]$.
\end{lemma}
\begin{proof}
	Fix $p \in [\Bar{p}]$. Let $k' \in \argmax_{k \in {\cal A}_p} \{ \overline{U}_k (p) \}$ represent the empirically best arm at the end of phase $p$. For the globally optimal $k^*$ to be eliminated by $k'$, we need to have $\overline{U}_{k^*} (p) + E_p < \overline{U}_{k'} (p) - E_p$ by \eqref{eqn:s3ae}. Since on ${\cal E}_{g}$, we have $\abs{\overline{U}_k(p) - \mu_k} \leq E_p$ for all $k \in \mathcal{A}_p$, then
		$\mu_{k^*} \leq \overline{U}_{k^*} (p) + E_p  
		\leq \overline{U}_{k'} (p) - E_p 
		\leq \mu_{k'}$,
	which is a contradiction since we have $\mu_{k^*} > \mu_{k'}$. 
\end{proof}

We utilize Lemma~\ref{lemma:globalremains} to prove the elimination of $\beta$-suboptimal arms in the following lemmas. In the following lemma, we will show that an arm in $\neg {\cal K}_{\beta}$ will be certainly eliminated under the good event when the number of samples $s(p)$ collected from that arm by the end of phase $p$ exceeds a suboptimality gap dependent threshold.

\begin{lemma}($\beta$-suboptimal arm elimination)\label{lemma:elimination}
	Under ${\cal E}_{g}$ and Assumption \ref{ass:discern}, 
    any $\beta$-suboptimal arm $k \in \neg \mathcal{K}_{\beta}$ will be eliminated by the end of phase $p_k$, where $p_k$ is the smallest integer satisfying
	\begin{align} \label{eqn:ia1}
		s(p_k) \geq \frac{\Lambda_1 \sigma^2}{(\Delta_k - \omega)^2}~.
	\end{align}
\end{lemma}
\begin{proof}
	First realize that by Lemmas \ref{corr:pt} and \ref{lemma:confidence_bound}, ${\cal E}_{g}$ holds for all phases relevant until elimination. Consider a $\beta$-suboptimal arm $k$. For such an arm, by the definitions of $E_p$ and $\omega$, \eqref{eqn:ia1} implies $4 E_{p_k} \leq \Delta_k$. We have
	\begin{align}
		\overline{U}_k(p_k) + E_{p_k} &\leq \mu_k + 2 E_{p_k} \label{eqn:fli4} \\
		&\leq \mu_k + 2E_{p_k} +  \overline{U}_{k^*}(p_k) - \mu_{k^*} +  E_{p_k} \label{eqn:fli5} \\ 
		&=  \overline{U}_{k^*}(p_k) + 3 E_{p_k} -\Delta_k \label{eqn:fli6} \\ 
		&\leq \overline{U}_{k^*}(p_k) - E_{p_k}~, \label{eqn:fli7}
	\end{align}
	where \eqref{eqn:fli4} and \eqref{eqn:fli5} hold under ${\cal E}_{g}$. \eqref{eqn:fli6} follows from the definition of the suboptimality gap $\Delta_k = \mu_{k^*} - \mu_k$, and \eqref{eqn:fli7} holds since we have $4 E_{p_k} \leq \Delta_k$. Finally, by the rule in \eqref{eqn:s3ae} and Lemma \ref{lemma:globalremains}, \eqref{eqn:fli7} then implies that a $\beta$-suboptimal arm $k$ will be eliminated by the end of phase $p_k$. 
\end{proof}

Lemma \ref{lemma:elimination} implies that arms that have a suboptimality gap $\Delta_k$ which is closer to the discernibility margin requires more samples to eliminate. The following lemma shows that all $\beta$-suboptimal arms are eliminated by termination phase.
\begin{lemma}(Termination)\label{lemma:termination}
	Under ${\cal E}_{g}$ and Assumption \ref{ass:discern} all of the $\beta$-suboptimal arms are eliminated by the time Fed-MoM-UCB terminates. 
\end{lemma}
\begin{proof}
	Recall that $p_{\tau}$ is the phase after which Fed-MoM-UCB terminates, that is, either $E_{p_{\tau}} \leq \beta / 4$ or $\abs{{\cal A}_{p_{\tau}}} = 1$ by Algorithm~\ref{alg:algo1}. If the latter is the case, we have ${\cal A}_{p_{\tau}} = \{k^*\}$ by Lemma~\ref{lemma:globalremains} and we are done. For the former case where $E_{p_{\tau}} \leq \beta / 4$, remember that we have $\Delta_k > \beta$ for a $\beta$-suboptimal arm $k \in \neg {\cal K}_{\beta}$. Then, similar to the proof of Lemma~\ref{lemma:elimination}, it immediately follows that,
	\begin{align}
		\overline{U}_k(p_{\tau}) + E_{p_{\tau}} &\leq \overline{U}_{k^*}(p_{\tau}) + 3 E_{p_{\tau}} -\Delta_k \label{eqn:fli8} \\ 
		&\leq \overline{U}_{k^*}(p_{\tau}) - E_{p_{\tau}}~, \label{eqn:fli9}
	\end{align}
	where \eqref{eqn:fli8} follows from \eqref{eqn:fli6}, and \eqref{eqn:fli9} follows from $\Delta_k > \beta \geq 4 E_{p_{\tau}}$, suggesting that the $\beta$-suboptimal is bound to be eliminated by the end of phase $p_{\tau}$.
\end{proof}

Lemmas \ref{lemma:globalremains} and \ref{lemma:termination} establish that Fed-MoM-UCB is a $(\beta,\delta)$-PAC algorithm. Next, we use the lemmas above to explicitly bound the sample complexity and the regret. 

\begin{theorem}(Sample complexity bound)\label{thm:sample_complexity}
    Under Assumption \ref{ass:discern}, the number of samples Fed-MoM-UCB requires to terminate with a set of $(\beta, \delta)$-PAC arms remaining is bounded above by
    \begin{align*}
        &M\sum_{k \in \neg {\cal K}_\beta} \left(\frac{\Lambda_1 \sigma^2}{(\Delta_k - \omega)^2} + l(p_k) \right) + M\sum_{k \in {\cal K}_\beta} \left( \frac{\Lambda_1 \sigma^2}{\epsilon^2} + l(p_\tau) \right)
    \end{align*}
\end{theorem}
\begin{proof}
    For the sample complexity associated with $\beta$-suboptimal arms, the proof draws on the results of Lemma \ref{lemma:elimination} and notes that in each phase $p$, each arm is sampled $l(p)$ times. Specifically, by \eqref{eqn:ia1}, we can write 
        $s(p_k) < \frac{\Lambda_1 \sigma^2}{(\Delta_k - \omega)^2} + l(p_k)$.    
    Regarding the sample complexity for $\beta$-optimal arms, we account for a sufficient number of rounds leading up to the termination of Fed-MoM-UCB. By Lemma \ref{corr:pmax}, we have
        $s(p_{\tau}) < \frac{\Lambda_1 \sigma^2}{\epsilon^2} + l(p_{\tau})$.
    To conclude the proof we sum the samples over all arms and clients.
\end{proof}

Next, we give a bound on the robust federated regret. Since Fed-MoM-UCB incurs no regret after the elimination of all $\beta$-suboptimal arms, we analyze the regret of our algorithm at termination, denoted $\Tilde{R}^\beta_\tau$.

\begin{theorem} \label{thm:regret_bound}
    Under Assumption \ref{ass:discern}, with probability at least $1-\delta$, the robust federated regret of Fed-MoM-UCB when it terminates is upper bounded as
 	\begin{align}
		\Tilde{R}^\beta_{\tau} < M \hspace{-0.1in} \sum_{k \in \neg \mathcal{K}_\beta}  \left( \frac{\Lambda_1 \sigma^2}{\Delta_k - \omega} + l(p_k) (\Delta_k - \beta) \right)  
		+ M  C p_{\tau} . \label{eqn:mainregret2}
	\end{align}
\end{theorem}

\begin{proof}
	Given that a $\beta$-suboptimal arm $k \in \neg {\cal K}_{\beta}$ is bound to be eliminated by the end of the phase $p_k$, we have
	\begin{align} \label{eqn:mainregret}
		\Tilde{R}^\beta_{\tau} \leq \sum_{k \in \neg \mathcal{K}_\beta} M (\Delta_k - \beta)  s(p_k) + CMp_{\tau}~.
	\end{align}
	By Lemma \ref{lemma:elimination} and the fact that  $s(p_k) < \frac{\Lambda_1 \sigma^2}{(\Delta_k - \omega)^2} + l(p_k)$, we have
	\begin{align}
		(\Delta_k - \beta) s(p_k) &< (\Delta_k - \beta) \left(\frac{\Lambda_1 \sigma^2}{(\Delta_k - \omega)^2} +  l(p_k)\right) \label{eqn:fli15} \\
		&< \frac{\Lambda_1 \sigma^2}{\Delta_k - \omega} + l(p_k) (\Delta_k - \beta)~, \label{eqn:fli13}
	\end{align}
	where \eqref{eqn:fli13} holds since $\beta > \omega$. 
	Theorem~\ref{thm:regret_bound} then follows after combining \eqref{eqn:mainregret} and \eqref{eqn:fli13}.
\end{proof}

Note that Theorems \ref{thm:sample_complexity} and \ref{thm:regret_bound} are given for generic phase lengths $l(p)$ and that $p_\tau \leq \bar{p}$ is a random variable contingent on the arm outcomes. In the regret bound,  $\Lambda_1 \sigma^2/(\Delta_k - \omega)$ term suggests that a smaller suboptimality gap $\Delta_k$ increases the regret as the suboptimal arm $k$ will remain active for a long time. $l(p_k)$ term multiplies $\Delta_k - \beta$ term (i.e., instantaneous regret), meaning longer phase lengths increase the regret. This is due to the fact that arms are eliminated only at the end of the phases. On the other hand, communication regret depends on $l$ through $p_{\tau}$. Therefore, smaller phase lengths results in an increase in the communication regret. Let $\tau$ denote the last round of phase $p_\tau$, which is the last round before termination. Since Fed-MOM-UCB will not incur any regret after the termination phase, the regret bound in Theorem \ref{thm:regret_bound} holds for all $T \geq \tau$, i.e., $R^{\beta}_T = \tilde{R}^{\beta}_{\tau}$. This theorem shows that the regret is $O(1)$ with at least $1-\delta$ probability in terms of $T$ dependence. The reason that we are able to get $O(1)$ instead of $O(\log T)$ regret is due to the facts that (i) our bound holds with probability $1-\delta$, (ii) we are content with $\beta$-optimal arms, which is a satisficing objective \cite{huyuk2021multi}. Since communication cost is a part of the regret, this result shows that the communication cost is also $O(1)$. However, the communication cost until termination can grow linearly depending on the choice of $l$. This is evident when the phase length is fixed, i.e., $l(p) = \kappa$ for $\kappa \in \mathbb{N}_{+}$. Different choices of $l$ will yield different tradeoffs between regret due to selecting suboptimal arms and regret due to communication. Two interesting cases are presented in the following corollaries.

\begin{corollary} \label{cor:2^p}
    When $l(1) = 2$, $l(p) = 2^{p-1}$ for $p>1$, i.e., $s(p) = 2^p$, under Assumption \ref{ass:discern}, with probability at least $1-\delta$, the number of samples Fed-MoM-UCB gathers before termination is bounded by 
    \begin{align*}
        &2M\sum_{k \in \neg {\cal K}_\beta} \frac{\Lambda_1 \sigma^2}{(\Delta_k - \omega)^2} + 2 M |{\cal K}_\beta| \frac{\Lambda_1 \sigma^2}{\epsilon^2}~,
    \end{align*}
    and the robust federated regret is bounded by
    \begin{align*}
        \Tilde{R}^\beta_{\tau} &< 2 M \sum_{k \in \neg \mathcal{K}_\beta}  \frac{\Lambda_1 \sigma^2}{\Delta_k - \omega}  + M C  \log_2 \left( \frac{\Lambda_1 \sigma^2}{\epsilon^2} \right) + MC  ~.
    \end{align*}
\end{corollary}

\begin{proof}
For $k \in \neg {\cal K}_{\beta}$, from the definition of $s(p_k)$ we have, 
\begin{align*}
    s(p_k-1) < \frac{\Lambda_1 \sigma^2}{(\Delta_k - \omega)^2} \Leftrightarrow p_k < \log_2 \left( \frac{\Lambda_1 \sigma^2}{(\Delta_k - \omega)^2} \right) + 1.
\end{align*}
Similarly,
\begin{align*}
    s(\bar{p}-1) < \frac{\Lambda_1 \sigma^2}{\epsilon^2} \Leftrightarrow \bar{p} < \log_2 \left( \frac{\Lambda_1 \sigma^2}{\epsilon^2} \right) + 1 .
\end{align*}
Substituting the above expressions in statement of Theorems \ref{thm:sample_complexity} and \ref{thm:regret_bound} yields the sample complexity and regret upper bounds in the theorem statement. 
\end{proof}

Corollary \ref{cor:2^p} shows that with geometrically increasing phase lengths, the regret due to communication will have logarithmic dependence on $1/\epsilon$, while the regret due to suboptimal arm selections increases only by a multiplicative factor. 

	\begin{corollary} \label{cor:ic1}
		In Fed-MOM-UCB, all $\beta$-suboptimal arms can be eliminated in a single communication round by setting $l(p) = \left\lceil \Lambda_1 \sigma^2 / \epsilon^2 \right\rceil$. Then, with probability at least $1-\delta$, the number of samples Fed-MoM-UCB gathers before termination is bounded by $M K \left\lceil \Lambda_1 \sigma^2 / \epsilon^2 \right\rceil$,
        and the robust federated regret is bounded by
			$M \sum_{k \in \neg {\cal K}_{\beta}} (\Delta_k - \beta) \left\lceil \Lambda_1 \sigma^2 / \epsilon^2 \right\rceil + C M$.
	\end{corollary}
	\begin{proof}
	Proceeding similar to the proof of Theorems~\ref{thm:sample_complexity} and \ref{thm:regret_bound}.
	\begin{align*}
		& s(1) = \left\lceil \frac{\Lambda_1 \sigma^2}{\epsilon^2} \right\rceil \geq \frac{\Lambda_1 \sigma^2}{\epsilon^2} 
		= \frac{64 \sigma^2}{\epsilon^2} \left(\frac{4\eta(\lambda)-1}{2\eta(\lambda)-1} \frac{\log 2}{B}\right) \\
		&\Rightarrow  \frac{2 \sigma}{\sqrt{s(1)}}  \left(\frac{4\eta(\lambda)-1}{2\eta(\lambda)-1} \frac{\log 2}{B}\right)^{1/2} \leq \frac{\epsilon}{4} 
		\Rightarrow E_1 \leq \frac{\omega + \epsilon}{4} = \frac{\beta}{4} 
	\end{align*}
	hence Fed-MOM-UCB terminates at the end of the first phase. 
	\end{proof}

	 Comparing Theorem~\ref{thm:regret_bound} with Corollary~\ref{cor:ic1}, we observe that the main difference is the replacement of the $O( (\Delta_k - \beta) / (\Delta_k - \omega)^2 )$ term in \eqref{eqn:fli15} with $O( (\Delta_k - \beta) / \epsilon^2)$. Even though a single round of communication is enough, arms with gaps greater than $\beta$ can be discarded after much less than $\lceil \Lambda_1 \sigma^2 /\epsilon^2 \rceil$ samples in multiple phases with smaller phase lengths.

\section{Experimental Results} \label{sec:experimentsmain}

In this section, we report our empirical observations regarding Fed-MoM-UCB. We test our algorithm against two benchmark algorithms. One of them is Fed2-UCB, proposed in \cite{shi2021federated}, which is specifically developed for federated bandits with non-i.i.d. data distributions. However, Fed2-UCB does not address the challenge of adversarial attacks. For this reason we present another robust benchmark we call Fed-Trim-UCB. Fed-Trim-UCB is a version of the trimmed mean based algorithm in \cite{neurips_bandit}, repurposed for the federated setting. Similar trimmed mean algorithms can be also found in \cite{trimmed, liu2023flgqm}. In particular, Fed-Trim-UCB uses the following estimator,
\begin{align*}
\overline{U}_{k,\text{trim}} (p) = \frac{1}{(1-2\lambda)M} \sum_{i = \lceil \lambda M \rceil}^{\lfloor(1-\lambda) M\rfloor} U_{(i), k} (p) ~,
\end{align*}
where $U_{(i), k} (p)$ is the order statistics of $U_{m,k} (p)$ with respect to the clients for a fixed arm $k$. We set the exploration threshold as suggested by \cite{neurips_bandit} to
\begin{align*}
E_{p, \text{trim}} = \frac{\varrho}{1-\lambda} \sqrt{\frac{2}{s(p)} \log \left( \frac{K s(p)^2 \pi^2}{12 \delta} \right)} ~,
\end{align*}
where arms are $\varrho$-subgaussian. For our setting we pick $\varrho = \sigma + \sigma_c$. Similar to Fed-MoM-UCB, we set the termination condition for Fed-Trim-UCB to $E_{p,\text{trim}} \leq \beta / 4$ and use the same client recruiting policy. For our analysis, we conducted simulations on a synthetic MAB problem, applying various attack schemes to evaluate the efficacy of Fed-MoM-UCB, Fed-Trim-UCB, and Fed2-UCB. We detail our findings across three distinct scenarios, with each scenario's objectives and results thoroughly discussed in its designated section. For a concise overview of these scenarios, see Table~\ref{tab:Scenarios}. The model and arm sampling noises are defined as zero-mean additive i.i.d. Gaussian noise with variance $\sigma_c^2 = 0.02^2$ and $\sigma^2 = 2^2$ for all scenarios. We use $\delta = 0.001$ and perform 100 independent experiments and report the mean and standard deviations for the cumulative regret and the communication cost. We calculate the cumulative regret using \eqref{eqn:regret2} for all of the algorithms, where the $\beta$-optimal arms do not incur any regret. 

\begin{table}[h]
	\centering
	\setlength{\tabcolsep}{4pt}
	\caption{Unknown global mean arm outcomes}
	\label{tab:armmeans}
	\begin{tabular}{@{}lccccccccc@{}}
		\toprule
		\multicolumn{1}{c}{$k$} & 1 & 2 & 3 & ... & 37 & 38 & 39 & 40 \\ \midrule
		$\mu_k$ & 0.00 & 0.05 & 0.10 & ... & 1.80 & 1.85 & 1.90 & 1.95 \\ \bottomrule
	\end{tabular}
\end{table}

\begin{table*}[t]
	\centering
	\caption{Parameters for different Scenarios. $M$, $B$, and $G$ columns show the number of total clients recruited, the number of clients in a group, and the number of groups for the Fed-MoM-UCB algorithm, respectively.}
	\label{tab:Scenarios}
	\tabcolsep=0.11cm
	\begin{tabular}{@{}lccccccclccclccccccccc@{}}
		\toprule
		\textbf{}  & & & & & & & & & \multicolumn{4}{c}{Fed-MoM-UCB} & & \multicolumn{2}{c}{Fed2-UCB} & & \multicolumn{2}{c}{Fed-Trim-UCB} \\ \cmidrule(lr){10-13} \cmidrule(lr){15-16} \cmidrule(l){18-20} 
		\textbf{} & $T$ & $\lambda$ & $\omega$ & $\beta$ & $\neg {\cal K}_{\beta}$ & $C$ & Attack &  & $\alpha(\lambda)$  & $M$ & $B$ & $G$ & & $M(p)$ & $M$  & & $M(p)$ & $M$                                               \\ \cmidrule(r){1-8} \cmidrule(lr){10-13} \cmidrule(l){15-16} \cmidrule(l){18-20}
		\multirow{2}{*}{\textbf{Scenario 1}} & \multirow{2}{*}{$5 \times 10^4$} & \multirow{2}{*}{0.05} & \multirow{2}{*}{0.143}         & \multirow{2}{*}{0.32} & \multirow{2}{*}{$\{1,\ldots,33\}$} & \multirow{2}{*}{1}   & \multirow{2}{*}{\eqref{eqn:sce1attack}} &  & \multirow{2}{*}{Geometric} & \multirow{2}{*}{126} & \multirow{2}{*}{3} & \multirow{2}{*}{42} & \multirow{2}{*}{} & $10p$                  & $1752 \pm 174$               & & \multirow{2}{*}{126} & \multirow{2}{*}{126}  \\
		&                                  &                      &                                &                        &                                   &                      &                                         &  &                             &                     &                    &                     &                   & $2^p$               & $12593 \pm 1452$        \\ \cmidrule(r){1-8} \cmidrule(lr){10-13} \cmidrule(l){15-16} \cmidrule(l){18-20} 
		\multirow{2}{*}{\textbf{Scenario 2}} & \multirow{2}{*}{$10^5$}          & 0.25                  & 0.298                          & 0.47                  & \multirow{2}{*}{$\{1,\ldots,30\}$} & \multirow{2}{*}{0.1} & \multirow{2}{*}{\eqref{eqn:sce2attack}} &  & \multirow{2}{*}{Arithmetic} & 65                  & 1                  & 65                  &                   & \multirow{2}{*}{20$p$} & $2796 \pm 176$  & & 65 & 65              \\
		&                                  & 0.3                 & 0.326                          & 0.47                  &                                   &                      &                                         &  &                             & 74                  & 1                  & 74                  &                   &                     & $1162 \pm 86$ & & 74 & 74              \\ \cmidrule(r){1-8} \cmidrule(lr){10-13} \cmidrule(l){15-16} \cmidrule(l){18-20} 
		\multirow{4}{*}{\textbf{Scenario 3}} & \multirow{4}{*}{$2 \times 10^5$} & \multirow{4}{*}{0.1} & $0.249$ & \multirow{4}{*}{0.48}  & \multirow{4}{*}{$\{1,\ldots,30\}$} & \multirow{4}{*}{2} & \multirow{4}{*}{\eqref{eqn:sce1attack}} &  & Arithmetic                   & 44                 & 1                  & 44                  &                   & \multirow{4}{*}{-} & \multirow{4}{*}{-}  & & \multirow{4}{*}{-} & \multirow{4}{*}{-}\\
		&                                  &                   &              0.184              &                        &                                   &                      &                                         &  & Geometric                    & 92                 & 2                  & 46                  &                   &                     &                               & & & &\\
		&                                  &                      &                 0.200              &                        &                                   &                      &                                         &  & Harmonic                  & 108                 & 2                  & 54                  &                   &                     &                               \\ 
      &                                  &                      &                 0.188              &                        &                                   &                      &                                         &  & Polynomial                  & 288                 & 4                  & 72                  &                   &                     &              \\                 
		\bottomrule
	\end{tabular}

\end{table*}
\begin{figure*}[t]
	\centering
	\begin{subfigure}[t]{0.24\textwidth}
		\centering
		\captionsetup{width=.9\linewidth}
		\includegraphics[width=\textwidth]{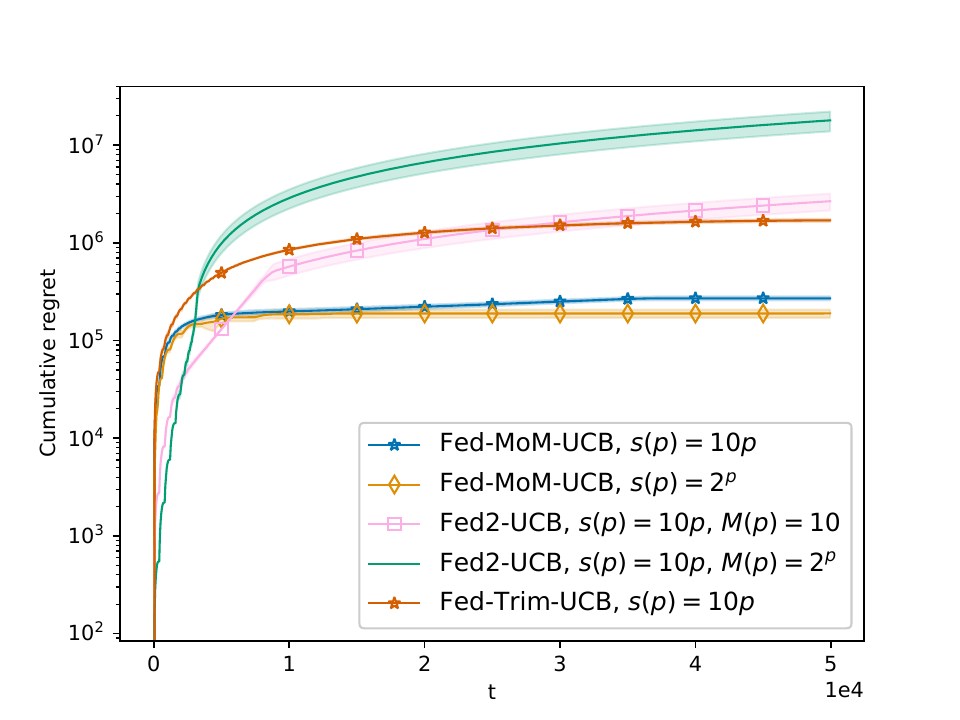}
		\caption{Cumulative Regrets}
		\label{fig:sce1_reg}
	\end{subfigure}%
	\begin{subfigure}[t]{0.24\textwidth}
		\centering
		\captionsetup{width=.75\linewidth}
		\includegraphics[width=\textwidth]{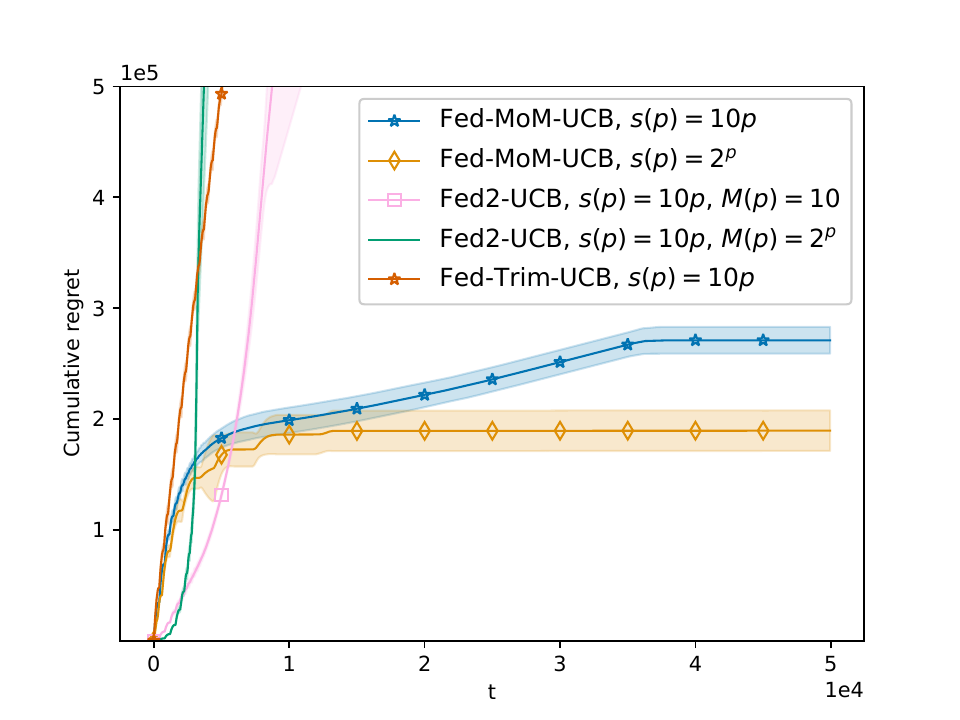}
		\caption{Cumulative Regrets (zoomed)}
		\label{fig:sce1_reg_zoom}
	\end{subfigure}%
	\begin{subfigure}[t]{0.24\textwidth}
		\centering
		\captionsetup{width=.85\linewidth}
		\includegraphics[width=\textwidth]{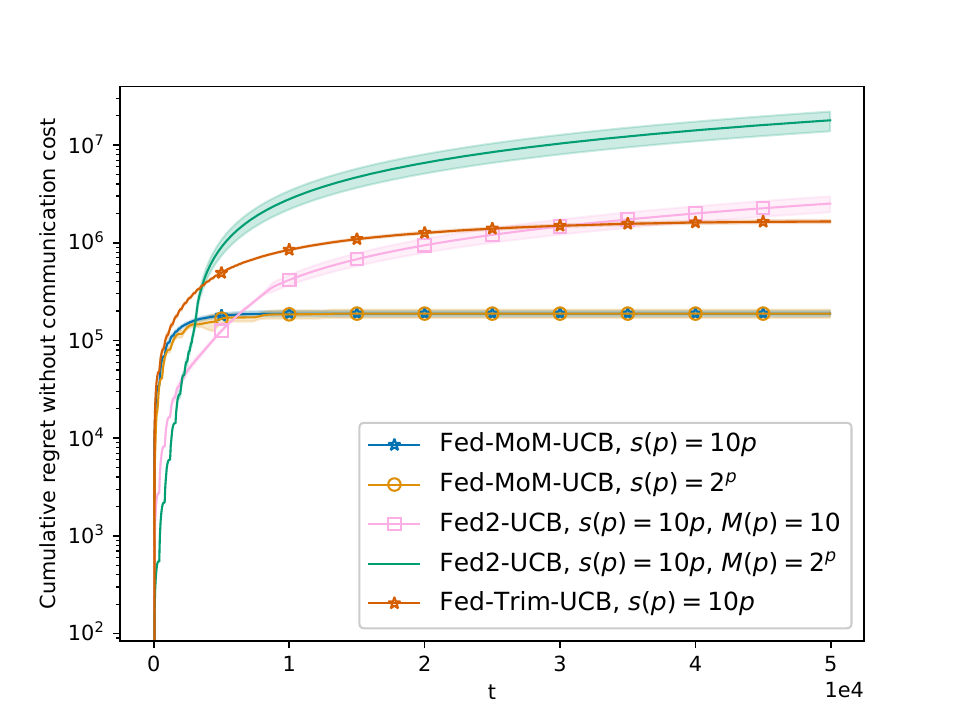}
		\caption{Cumulative regrets without communication costs}
		\label{fig:sce1_nocomm_reg}
	\end{subfigure}
	\begin{subfigure}[t]{0.24\textwidth}
		\centering
		\captionsetup{width=.9\linewidth}
		\includegraphics[width=\textwidth]{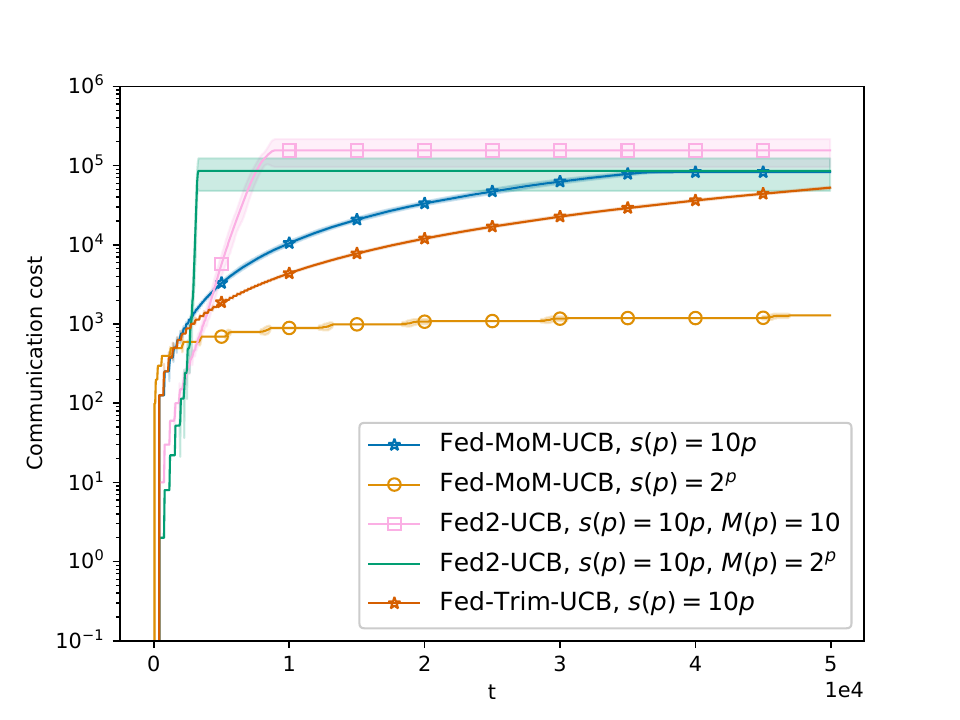}
		\caption{Communication costs only}
		\label{fig:sce1_comm_reg}
	\end{subfigure}
	\caption{Regrets and communication costs for Scenario 1.}
	\label{fig:sce1}
\end{figure*}

\subsection{Scenario 1} \label{ssec:sce1}
We choose $\beta = 0.32$ which gives the $\beta$-suboptimal arm set as $\neg {\cal K}_{\beta} = \{1,\ldots,33\}$. For this scenario we consider a case where the fraction of Byzantine clients is small but the attacking scheme is strong. We set the Byzantine clients' fraction to $\lambda = 0.05$ and use the geometric mapping for Fed-MoM-UCB, that is, $\alpha(\lambda)=\sqrt{2\lambda}$ (see Table~\ref{tab:tau}). We choose the geometric mapping as it yields a small value for the discernibility margin at $\lambda = 0.05$, which is $\omega = 0.149$ while also resulting in small constants $\Lambda_1$ and $\Lambda_2$ that scale the number of clients to be recruited and the cumulative regret. The communication cost is set to $C=1$. Finally, we consider the following outlier attack scheme for a Byzantine client $m$,
\begin{equation} \label{eqn:sce1attack}
	X_{m,k} (t)= \begin{dcases}
		\mathcal{N}(-14, 1) &\text{ if } k \in {\cal K}_{\beta} \\
		\mathcal{N}(14, 1) &\text{ if } k \in \neg {\cal K}_{\beta}
	\end{dcases}
\end{equation}
where $X_{m,k} (t)$ denotes the local observation of the Byzantine client $m$ for the arm $k$ at round $t$. That is, a Byzantine client $m$ samples its local observations from a normal distribution with mean $-14$ for the $\beta$-optimal arms and from a normal distribution with mean $14$ for the $\beta$-suboptimal arms, to make it harder for the server to identify the $\beta$-optimal arms correctly.
\begin{figure*}[t]
	\centering
	\begin{subfigure}[t]{0.24\textwidth}
		\centering
		\captionsetup{width=.9\linewidth}
		\includegraphics[width=\textwidth]{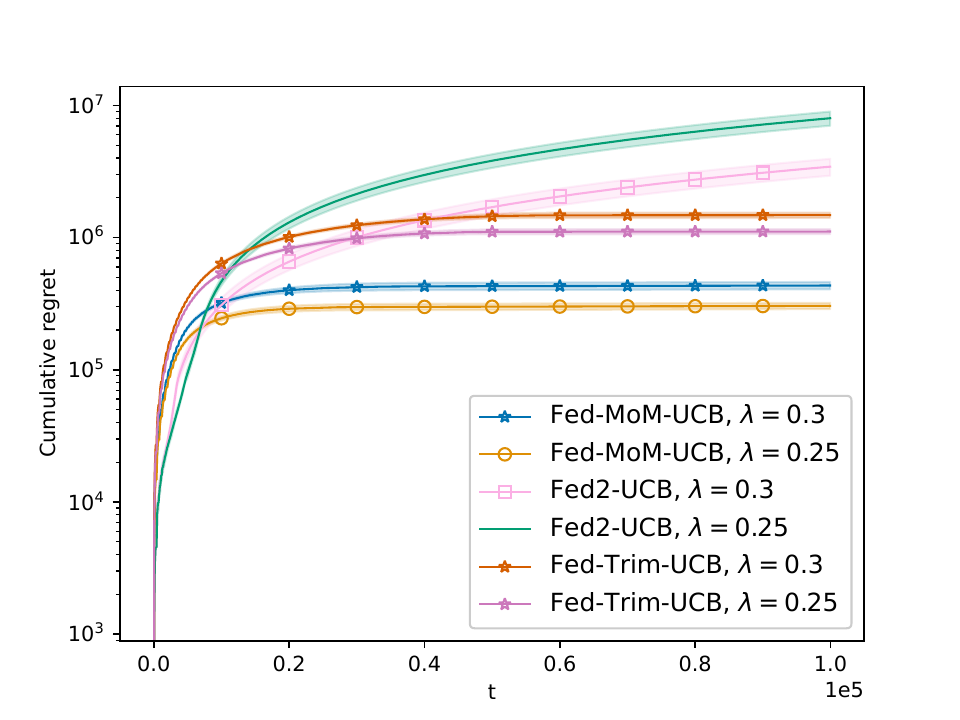}
		\caption{Cumulative Regrets}
		\label{fig:sce2_reg}
	\end{subfigure}%
	\begin{subfigure}[t]{0.24\textwidth}
		\centering
		\captionsetup{width=.75\linewidth}
		\includegraphics[width=\textwidth]{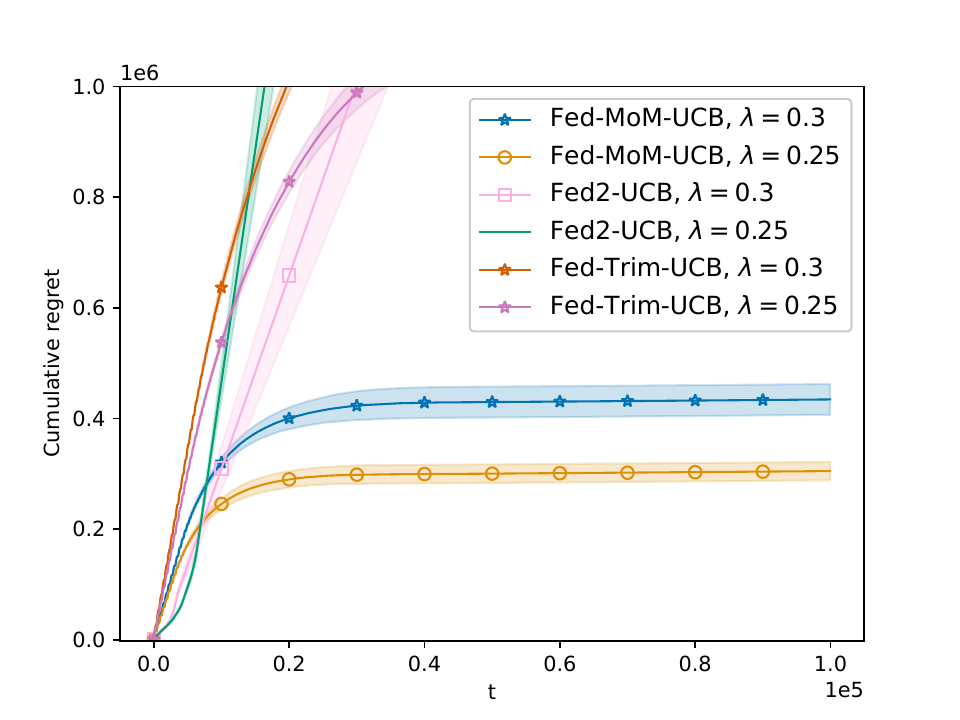}
		\caption{Cumulative Regrets (zoomed)}
		\label{fig:sce2_reg_zoom}
	\end{subfigure}%
	\begin{subfigure}[t]{0.24\textwidth}
		\centering
		\captionsetup{width=.85\linewidth}
		\includegraphics[width=\textwidth]{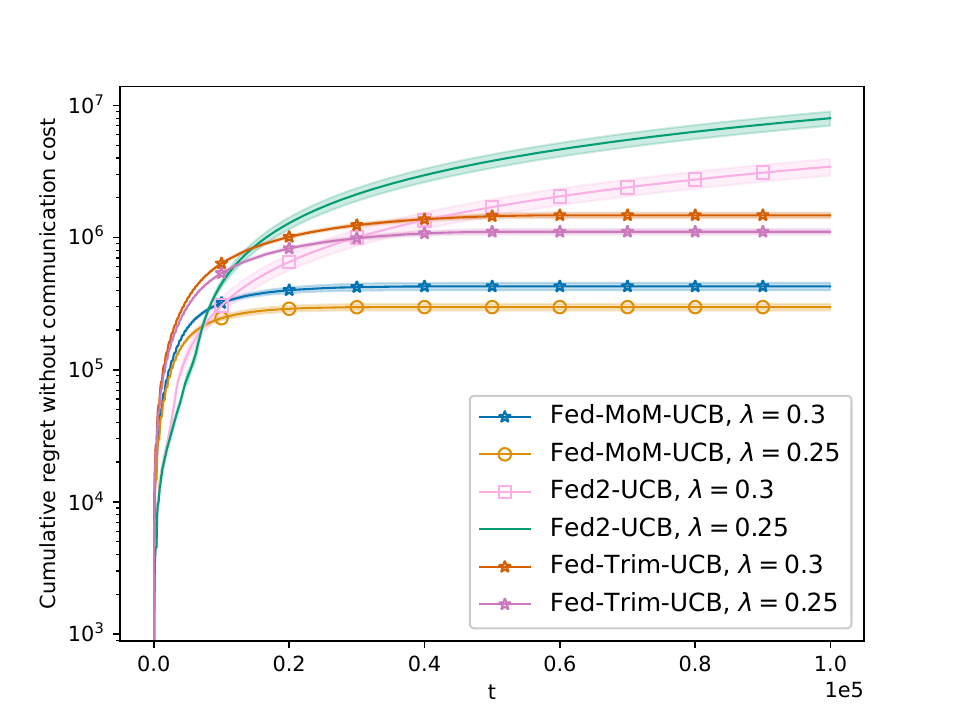}
		\caption{Cumulative regrets without communication costs}
		\label{fig:sce2_nocomm_reg}
	\end{subfigure}
	\begin{subfigure}[t]{0.24\textwidth}
		\centering
		\captionsetup{width=.9\linewidth}
		\includegraphics[width=\textwidth]{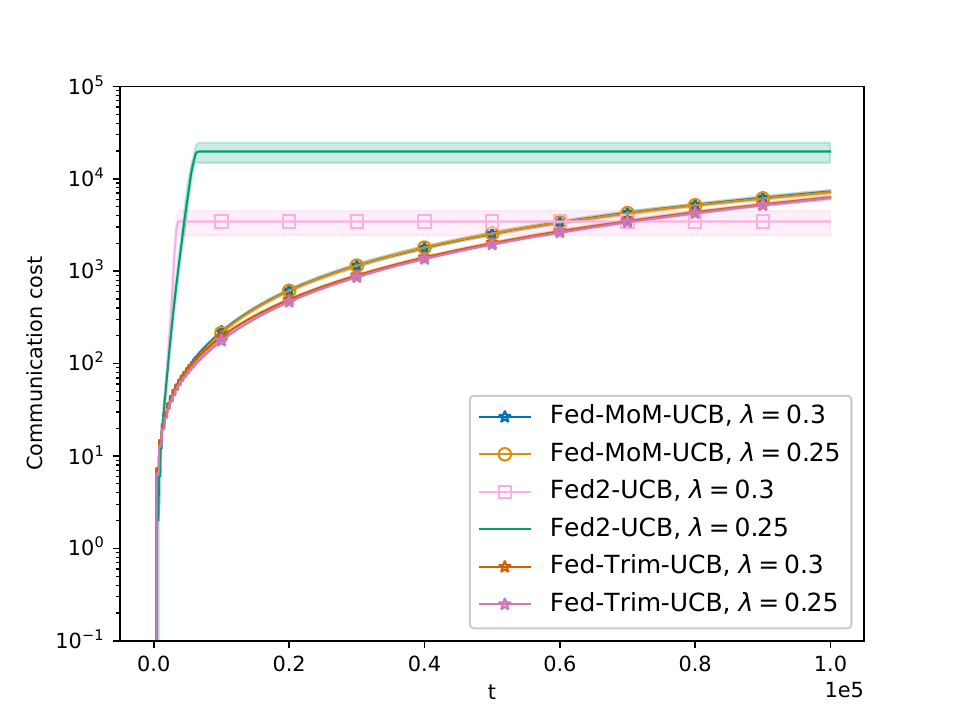}
		\caption{Communication costs only}
		\label{fig:sce2_comm_reg}
	\end{subfigure}
	\caption{Regrets and communication costs for Scenario 2.}
	\label{fig:sce2}
\end{figure*}
Unlike Fed-MoM-UCB, Fed2-UCB continues to recruit more clients over phases to reduce the uncertainty emerging from the client sampling. As discussed in Section~\ref{sec:regretanalysismain}, that does not help in the presence of Byzantine clients. We set $s(p) = 10p$ and run Fed2-UCB with two different configurations, $M(p) = 10p$ and $M(p) = 2^p$, where $M(p)$ is the total number of clients recruited at phase $p$. Fig.~\ref{fig:sce1_comm_reg} confirms that Fed2-UCB terminates. That is, only a single active arm is left, and the communication is stopped. However, the remaining arm is not a $\beta$-optimal arm and Fed2-UCB incurs linear regret in both cases as can be seen in Fig.~\ref{fig:sce1_reg} and Fig.~\ref{fig:sce1_reg_zoom}. Moreover, while recruiting more clients does not help Fed2-UCB eliminate the $\beta$-suboptimal arms, the cumulative regret also explodes since the more clients are recruited, the more the suboptimal arms are played. Table~\ref{tab:Scenarios} shows the extreme difference between the total number of clients recruited by Fed2-UCB and by Fed-MoM-UCB when they terminate. The unlimited client recruitment scheme of Fed2-UCB not only causes cumulative regret to increase significantly in the presence of Byzantine clients but may also be impractical for certain applications. In contrast, Fed-MoM-UCB achieves the $\beta$-optimal arm identification goal with a much smaller, fixed number of clients. In real-world applications, such as the treatment evaluation example in Section \ref{sec:motivating}, progressively recruiting additional clients may be unrealistic.

Fed-MoM-UCB recruits $M = 126$ only once at the beginning and creates $G=42$ groups. Note that even though it is possible to have $B\geq4$ as well while still maintaining the robustness, the geometric mapping is the best option for $\lambda=0.05$ since it gives the small $\omega$, $\Lambda_1$, and $\Lambda_2$ terms. We run Fed-MoM-UCB with two different configurations, $s(p) = 10p$ and $s(p) = 2^p$ as in Corollary \ref{cor:2^p}. Fig.~\ref{fig:sce1_comm_reg} confirms that Fed-MoM-UCB terminates and stops the communication in both cases. From Fig.~\ref{fig:sce1_reg_zoom}, we can see that Fed-MoM-UCB is able to eliminate all the $\beta$-suboptimal arms as the cumulative regret cease to increase. We run Fed-Trim-UCB with $s(p) = 10p$. Similar to Fed-MoM-UCB, Fed-Trim-UCB also achieves to eliminate all the $\beta$-suboptimal arms. However, as can be seen from Fig.~\ref{fig:sce1} the elimination requires more samples compared to Fed-MoM-UCB, hence the cumulative regret is greater. Also note that while Fed-Trim-UCB manages to eliminate all $\beta$-suboptimal arms, it fails to terminate in $5\times 10^4$ rounds, hence the communication cost is higher. This is because the confidence bound for Fed-Trim-UCB decreases slower compared to Fed-MoM-UCB.
\subsection{Scenario 2} \label{ssec:sce2}
The unknown global mean arm outcomes, model uncertainty variance and the sampling noise variance are the same as in Scenario 1. In this scenario we look at how a larger fraction of Byzantine clients effect the algorithms. We consider the following attack scheme for a Byzantine client $m$,
\begin{equation} \label{eqn:sce2attack}
	X_{m,k} (t)= \begin{dcases}
		\texttt{Unif}[-0.2,~-0.1] &\text{ if } k \in {\cal K}_{\beta} \\
		\mu_{m,k} + \zeta_{m,t} &\text{ if } k \in \neg {\cal K}_{\beta}
	\end{dcases}
\end{equation}
That is, a Byzantine client $m$ sets its local observations to a random value in $[-0.2,~-0.1]$ for the $\beta$-optimal arms, while taking and recording honest samples for the $\beta$-suboptimal arms (see \eqref{eqn:reward_eqn}). The attack in \eqref{eqn:sce2attack} is milder than the one in \eqref{eqn:sce1attack}.

We experiment with two different values for the fraction of Byzantine clients, setting $\lambda = 0.25$ and $\lambda = 0.3$. In both cases, we run Fed2-UCB with $s(p) = 10p$ and $M(p) = 20p$, Fed-MoM-UCB and Fed-Trim-UCB with $s(p) = 10p$. We also change the communication cost to $C = 0.1$ to provide a more nuanced comparison. Fig.~\ref{fig:sce2} shows that even though the time horizon for the simulation is not enough for Fed-MoM-UCB to terminate, it still manages to eliminate all $\beta$-suboptimal arms and outperforms both Fed2-UCB and Fed-Trim-UCB, regardless of whether communication loss is taken into account. Similar to the first scenario, Fed2-UCB is unable to eliminate all $\beta$-suboptimal arms, whereas both Fed-MoM-UCB and Fed-Trim-UCB succeed in this task. Moreover, Fed-MoM-UCB proves to be more efficient in eliminating suboptimal arms, accruing less regret.

 We remark that depending on the attack scheme and the fraction of the malicious clients, Fed2-UCB may continue to work well under Byzantine attacks. However, it can not guarantee sublinear regret under arbitrary attacks. On the other hand, Fed-MoM-UCB can eliminate all the $\beta$-suboptimal arms when $\lambda < 0.5$.

\subsection{Scenario 3} \label{ssec:sce3}
In this part, we look at how the different choices for the $\alpha(\lambda)$ mapping effects the performance. The model noise and the sampling noise as well as the global arm means are the same as those used in Scenario 1. We set the communication cost to $C=2$ and the Byzantine client fraction to $\lambda = 0.1$. The attack scheme of a Byzantine client is the same as Scenario 1, and it follows $\eqref{eqn:sce1attack}$.

\begin{figure}[t]
	\centering
	\begin{subfigure}[t]{0.49\columnwidth}
		\centering
		\captionsetup{width=.9\linewidth}
		\includegraphics[width=\textwidth]{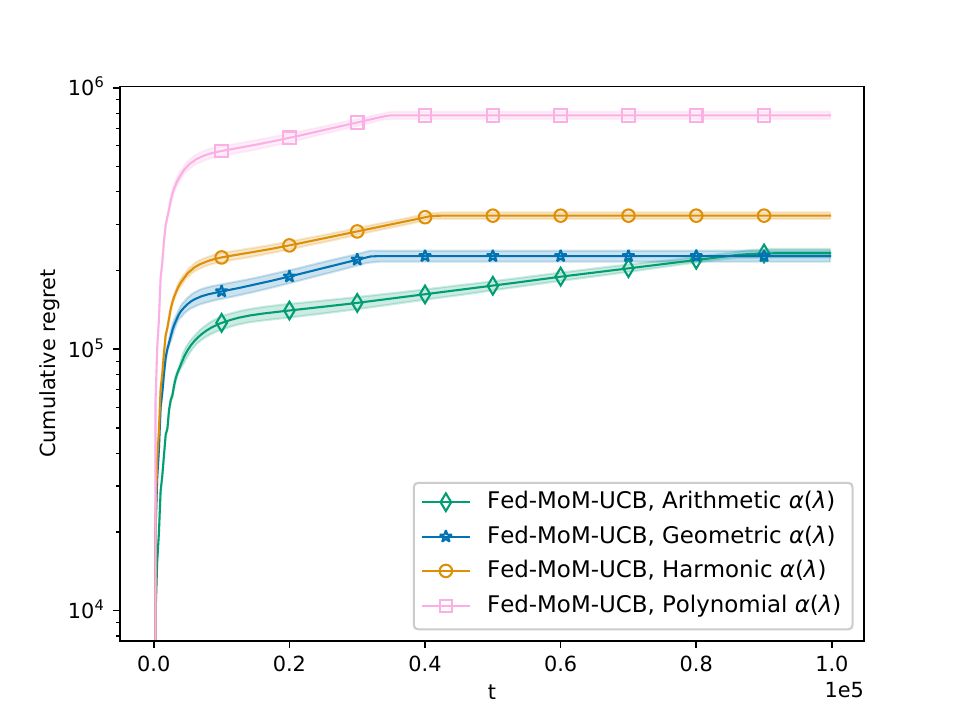}
		\caption{Cumulative Regrets}
		\label{fig:sce3_reg}
	\end{subfigure}%
	\begin{subfigure}[t]{0.49\columnwidth}
		\centering
		\captionsetup{width=.9\linewidth}
		\includegraphics[width=\textwidth]{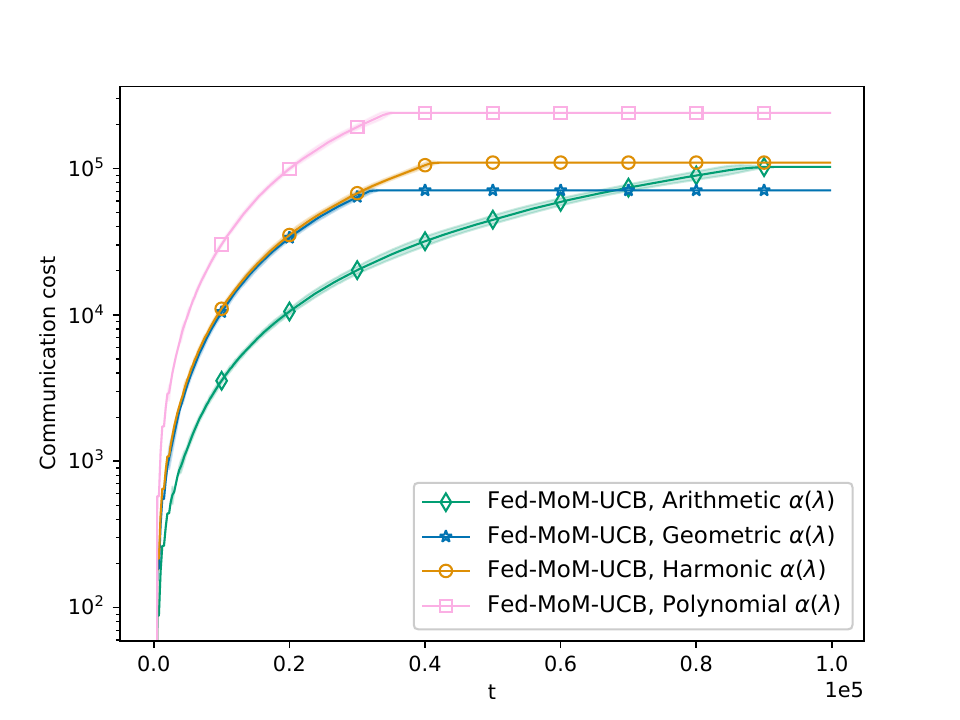}
		\caption{Communication Cost}
		\label{fig:sce3_comm}
	\end{subfigure}%
	\caption{Regrets and confidence bounds for Scenario 3.}
	\label{fig:sce3}
\end{figure}

Next, we analyze the mappings $\alpha(\lambda)$ given in Table~\ref{tab:tau}. Calculating $\omega$ for each choice, we observe that the smallest $\omega$ value is obtained with the geometric mapping which also accumulates the smallest regret and terminates the quickest. While due to its small number of client recruitment, the arithmetic mapping incurs small regret, it results in the largest number of rounds necessary to terminate. This is because while the small group size resulting from the arithmetic mapping gives rise to a small $M$, it also increases the number of samples required for elimination of $\beta$-suboptimal arms. Fig.~\ref{fig:sce3_reg} shows, the order between the cumulative regrets of the mappings as well as the communication costs. The geometric mapping offers the best performance with $M=92$ client, while the polynomial mapping offers the worst with the maximum number of clients $M=288$. Fig.~\ref{fig:sce3_comm} shows all mappings result in the termination of the algorithm with the geometric mapping giving the quickest termination and the arithmetic mapping giving the slowest.

\section{Conclusion and Future Work}

We focused on the problem of identifying $\beta$-optimal arms in a federated multi-armed bandit setting under model heterogeneity and Byzantine attacks, and proposed a robust median-of-means-based algorithm, Fed-MoM-UCB. We showed that when less than half of the clients are Byzantine, it is possible to maintain robustness while minimizing communication cost; thereby achieving low regret. We showed that our algorithm guarantees the elimination of arms with suboptimality gaps greater than a discernibility margin emerging from the size of the client groups, and gave $(\beta, \delta)$-PAC sample complexity results. We argued that recruiting more clients does not solve that problem, and we derived an expression for the required number of clients necessary to provide high probability guarantees on the regret and the sample complexity results. Finally, we evaluated the performance of Fed-MoM-UCB by comparing it with two other algorithms: one based on a trimmed mean approach, which we refer to as Fed-Trim-UCB, and the other known as Fed2-UCB. Our analysis confirmed the effectiveness of Fed-MoM-UCB. Interesting future work includes a detailed investigation on reducing the discernibility margin emanating from client sampling while retaining robustness against adversaries.

\section*{Appendix}

\subsection*{Proof of Lemma 1}

	The proof builds upon the proof of MoM concentration bound given in \cite[Proposition~2]{lafourge}.  Let ${\mathcal{G}}_p$ denote the set of client groups formed by the server at the end of phase $p$ and ${\mathcal{G}}^{\text{tru}}_p \subseteq {\mathcal{G}}_p$ denote the subset of groups with no Byzantine clients. We denote by $G_p := \abs{{\mathcal{G}}_p}$ and $G^{\text{tru}}_p := |{\mathcal{G}}^{\text{tru}}_p|$ the cardinalities of these sets. There is no concentration guarantee for the groups that include Byzantine clients. The worst case is when each Byzantine client belongs to a different group, rendering the maximum number of groups unreliable. Since $M^{\text{mal}} \leq \lambda M$ due to Assumption~\ref{ass:maliciousratio}, at most $\lfloor \lambda M \rfloor$ of the $G_p$ groups will be corrupted by Byzantine clients. Then, we have, 
	\begin{align}
		G^{\text{tru}}_p  \geq G_p - \lambda M \geq G_p - \frac{\lambda G_p}{\alpha(\lambda)} = G_p \eta(\lambda) ~. \label{eqn:momlemma2}
	\end{align}
	
	Fix $\gamma_p > 0$. By definition of the sample median, if the group mean of arm $k$ is within $\gamma_p$-neighborhood of $\mu_k$ for more than half of the groups, then the sample median of the group means must be within $\gamma_p$-neighborhood of $\mu_k$ as well. Thus, we have
	\begin{align}
		& \mathbb{P} (|\overline{U}_k (p) - \mu_k| > \gamma_p | {\mathcal{G}}_p)  \nonumber \\
		&\leq \mathbb{P} \left( \sum_{i \in {\mathcal{G}}^{\text{tru}}_p} \mathbb{I} \{ |U^i_k (p) - \mu_k | > \gamma_p \} \geq G^{\text{tru}}_p - \frac{G_p}{2}  \bigg| {\mathcal{G}}_p \right)  \nonumber \\ 
		&\leq \mathbb{P} \left( \sum_{i \in {\mathcal{G}}^{\text{tru}}_p} \mathbb{I} \{ |U^i_k (p) - \mu_k | > \gamma_p  \} \geq \dfrac{2\eta(\lambda) - 1}{2\eta(\lambda)} G^{\text{tru}}_p \bigg| {\mathcal{G}}_p  \right)  \nonumber \\
		&\leq\sum_{g = \lceil \frac{2\eta(\lambda) - 1}{2\eta(\lambda)}G_p^{\text{tru}} \rceil}^{G_p^{\text{tru}}} {G_p^{\text{tru}} \choose g} q_p^g (1-q_p)^{G_p^{\text{tru}} - g} \label{eqn:beforebinomialtrick} \\
		& \leq q_p^{ \frac{2\eta(\lambda) - 1}{2\eta(\lambda)}G_p^{\text{tru}} } 2^{G_p^{\text{tru}}} ~, \label{eqn:beforebinomialtrick2} 
	\end{align}
	where $q_p \coloneqq \mathbb{P}(|U^i_k(p) - \mu_k | > \gamma_p | {\mathcal{G}}_p )$. For \eqref{eqn:beforebinomialtrick}, observe that $\sum_{i \in {\mathcal{G}}^{\text{tru}}_p} \mathbb{I} \{ |U^i_k (p) - \mu_k | > \gamma_p  \}$ is a binomial random variable. Next, we observe the following for $i \in {\cal G}^{\text{tru}}_p$,
	\begin{align}
		U^i_k(p) - \mu_k  = \frac{1}{B} \sum_{m \in {\mathcal{B}}_{i,p}} ( \widehat{\mu}_{m,k} (p)  - \mu_{m,k} + \mu_{m,k} - \mu_k)  \notag \\
		= \underbrace{ \frac{1}{B} \sum_{m \in {\mathcal{B}}_{i,p}}  ( \widehat{\mu}_{m,k} (p)  - \mu_{m,k} ) }_{\text{Term I}}
		+ \underbrace{ \frac{1}{B} \sum_{m \in {\mathcal{B}}_{i,p}}  ( \mu_{m,k} - \mu_k ) }_{\text{Term II}} \notag ~.
	\end{align}
	By definition, Term I is the average of $B$ independent $\sigma/\sqrt{s(p)}$-subgaussian random variables. Thus, Term I is $\sigma/\sqrt{s(p)B}$-subgaussian. 
	Similarly, Term II is the average of $B$ independent $\sigma_c$-subgaussian random variables. Therefore, it is $\sigma_c/\sqrt{B}$-subgaussian. As a result, Term I plus Term II is $\sigma/\sqrt{s(p)B} + \sigma_c/\sqrt{B}$-subgaussian. Let $\rho \coloneqq \sigma/\sqrt{s(p)} + \sigma_c$ and $\rho' \coloneqq \frac{1}{\sqrt{B}} \rho$. Then, using the subgaussian tail inequality in \cite[Theorem~5.3 ]{lattimore_2020}), we have
	\begin{align}
		q_p & \leq 2 e^{- \gamma_p^2 / (2 (\rho')^2)} = 2e^{- B\frac{\gamma_p^2}{2 \rho^2}} \label{eqn:momlemma8} \\
		&\leq 2e^{- B\frac{4 \rho^2 \left(\frac{4\eta(\lambda)-1}{2\eta(\lambda)-1} \frac{\log 2}{B}\right) }{2 \rho^2}} \label{eqn:momlemma6}   \\
		&\leq \left( 0.5 \right)^{ \frac{2\eta(\lambda)}{2\eta(\lambda)-1}}  ~, \label{eqn:momlemma7}  
	\end{align}
	where \eqref{eqn:momlemma6} follows from \eqref{eqn:s4i1}, and \eqref{eqn:momlemma7} follows since $\eta(\lambda) > 1/2$ implies that $4\eta(\lambda) -1 > 2\eta(\lambda) $. From \eqref{eqn:beforebinomialtrick2}, observe that \eqref{eqn:momlemma7} holds and $\eta(\lambda)G_p - G_p^{\text{tru}} \leq 0$, which gives
	\begin{align}
		\eqref{eqn:beforebinomialtrick2} \leq q_p^{\frac{2 \eta(\lambda) - 1}{2}G_p} 2^{\eta(\lambda)G_p} ~. \label{eqn:momlemma9}
	\end{align}

	Finally, by plugging the tail bound for $q_p$ in \eqref{eqn:momlemma8} to the expression in \eqref{eqn:momlemma9}, we obtain 
	\begin{align*}
		\mathbb{P} (|\overline{U}_k (p) - \mu_k| > \gamma_p ) \leq \exp \left( - \frac{2 \eta(\lambda) -1}{16 \rho^2} M \gamma_p^2\right)~, 
	\end{align*}
	for all
	%
		$\gamma_p^2 \geq \frac{4 \rho^2}{B} \frac{4\eta(\lambda)-1}{2\eta(\lambda)-1} \log 2$,
	which completes the proof.

\subsection*{Proof of Lemma \ref{lemma:confidence_bound}}

For a given $\delta \in (0,1)$, by utilizing a union bound over $\Bar{p}$ phases and all active arms, we can bound $\mathbb{P} ( {\cal E}_{g} ) $ as
\begin{align}
		\mathbb{P}( {\cal E}_{g} )
        &= 1 - \mathbb{P} ( \exists p \in [\Bar{p}], \exists k \in {\cal A}_p : |\overline{U}_k (p) - \mu_k| > E_p ) \nonumber\\
        &= 1 - \mathbb{P} \left( \bigcup_{p \in [\Bar{p}]} \bigcup_{k \in {\cal A}_p} \left\{ |\overline{U}_k (p) - \mu_k| > E_p \right\} \right) \nonumber\\
		&\geq 1 - \sum_{p \in [\Bar{p}]} \sum_{k \in [K]} \mathbb{P} \left( |\overline{U}_k (p) - \mu_k| > E_p \right) ~, \label{eqn:lii2}
	\end{align}
where \eqref{eqn:lii2} follows from the union bound. To get a $1 - \delta$ bound on $\mathbb{P} ( {\cal E}_{g} ) $, we set the confidence in Lemma \ref{lemma:global} as
\begin{align}
    & \mathbb{P} \left( |\overline{U}_k (p) - \mu_k| > E_p \right) \leq \exp \left( - \frac{2 \eta(\lambda) -1}{16 \rho^2} M E_p^2\right) \leq \frac{\delta}{K \Bar{p}} \nonumber\\
    &\Leftrightarrow M \geq \frac{16 \rho^2 \log(  \frac{K \Bar{p}}{\delta} )}{(2 \eta(\lambda) -1) E^2_p } 
    \Leftrightarrow M \geq \Lambda_2  B \log(\frac{K \Bar{p}}{\delta}) ~, \label{eqn:lii3} 
\end{align} 
where \eqref{eqn:lii3} follows from the definitions of $\rho$ and $E_p$. Finally, observe that our choice of $M$ satisfies \eqref{eqn:lii3}.

\bibliography{main}
\bibliographystyle{IEEEtran}

\end{document}